\documentclass[lettersize,conference]{IEEEtran}

\usepackage{graphicx}
\usepackage{amsmath,amsfonts}
\usepackage{algorithmic}
\usepackage{algorithm}
\usepackage{array}
\usepackage{textcomp}
\usepackage{stfloats}
\usepackage{url}
\usepackage{verbatim}
\usepackage{cite}
\usepackage{ulem}

\newcommand{\lmm}[1]{\textcolor{green}{#1}}

\newcommand{\CUT}[1]{}
\newcommand{\ie}{\textit{i}.\textit{e}.}
\newcommand{\eg}{\textit{e}.\textit{g}.}

\usepackage{graphicx}

\usepackage{amsmath}
\usepackage{amssymb}
\usepackage{booktabs}
\usepackage{colortbl}
\usepackage{subcaption}
\usepackage{multirow}
\usepackage[pagebackref,breaklinks,colorlinks]{hyperref}
\definecolor{cssgreen}{rgb}{0.0, 0.5, 0.0}
\definecolor{cssred}{rgb}{1, 0, 0.0}
\definecolor{cssorange}{rgb}{1, 0.5, 0.0}

\hypersetup{
  colorlinks=true,
  citecolor=cssgreen,
  urlcolor=red,
}

\newenvironment{Itemize}%
{
\setlength{\leftmargini}{9pt}%
\begin{itemize}%
\setlength{\itemsep}{0pt}%
\setlength{\topsep}{0pt}%
\setlength{\partopsep}{0pt}%
\setlength{\parskip}{0pt}}%
{\end{itemize}}

\begin{document}

\title{E4S: Fine-grained Face Swapping via Editing With Regional GAN Inversion}

\definecolor{mygray}{gray}{.9}

\author{Maomao Li$^{1 *}$,
        Ge Yuan$^{1 *}$,
        Cairong Wang$^{2}$,
        Zhian Liu$^{3}$,
        Yong Zhang$^{4 \dagger}$,
        Yongwei Nie$^{3}$, \\
        Jue Wang, and 
        Dong Xu$^{1 \dagger}$,  \\
${^1}$ The University of Hong Kong \qquad
${^2}$Tsinghua Shenzhen International Graduate School \\
${^3}$ South China University of Technology 
\qquad 
${^4}$Tencent AI Lab \\       
}

\markboth{Journal of \LaTeX\ Class Files,~Vol.~14, No.~8, August~2021}%
{Shell \MakeLowercase{\textit{et al.}}: A Sample Article Using IEEEtran.cls for IEEE Journals}



\vspace{-1cm}
\twocolumn[{%
\renewcommand\twocolumn[1][]{#1}%
\maketitle
\begin{center}
    \centering
    \captionsetup{type=figure}
    \captionsetup{belowskip=-4pt}
    \includegraphics[width=\textwidth]{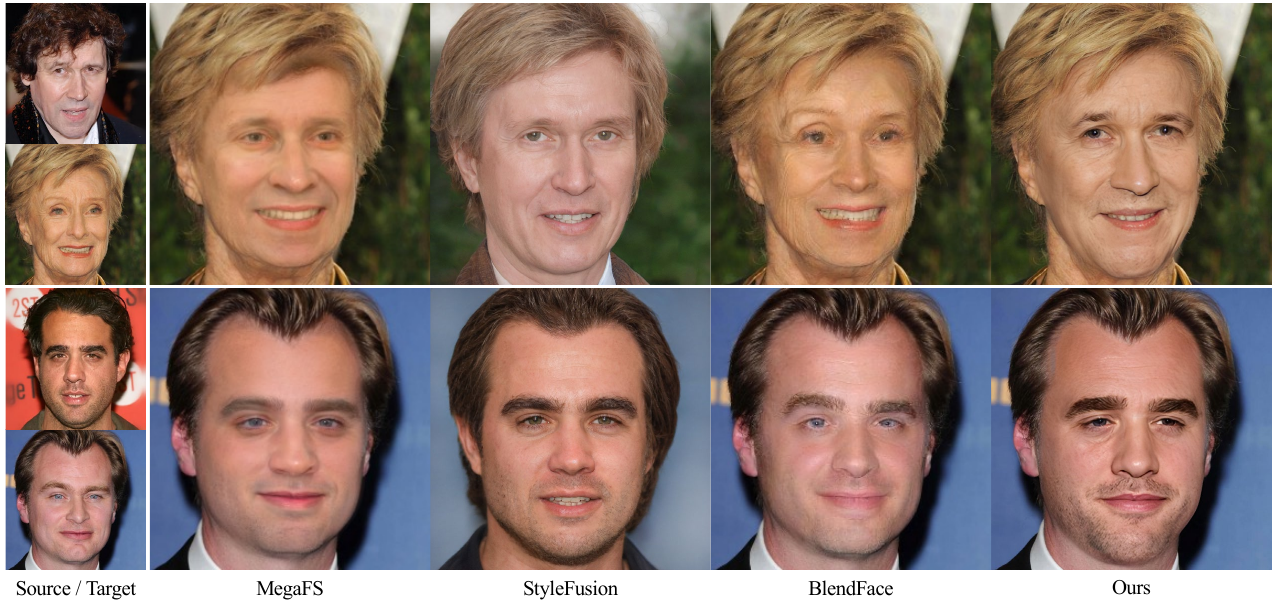}
    \vspace{-0.4cm}
       \caption{Compared with the existing StyleGAN-based face swapping approaches MegaFS~\cite{zhu2021MegaFS}, StyleFusion~\cite{kafri2021stylefusion}, and BlendFace~\cite{shiohara2023blendface}, our E4S can achieve high-fidelity swappped results. Note that ours achieves the best source identity and target attributes preservation, where the source skin tone and face shape are kept and the target lighting condition is well transferred to the swapped results. All the facial images except those of BlendFace~\cite{shiohara2023blendface} are at $1024\times1024$.} 
   
    \label{fig:teaser}
\end{center}%
}]

\begin{abstract}
This paper proposes a novel approach to face swapping from the perspective of fine-grained facial editing, dubbed \textit{``editing for swapping''} (E4S). The traditional face swapping methods rely on global feature extraction and fail to preserve the detailed source identity. In contrast, we propose a Regional GAN Inversion (RGI) method, which allows the explicit disentanglement of shape and texture. Specifically, our E4S performs face swapping in the latent space of a pretrained StyleGAN, where a multi-scale mask-guided encoder is applied to project the texture of each facial component into regional style codes and a mask-guided injection module manipulating feature maps with the style codes. Based on this disentanglement, face swapping can be simplified as style and mask swapping. Besides, due to the large lighting condition gap, transferring the source skin into the target image may lead to disharmony lighting. We propose a re-coloring network to make the swapped face maintain the target lighting condition while preserving the source skin. Further, to deal with the potential mismatch areas during mask exchange, we design a face inpainting module to refine the face shape. The extensive comparisons with state-of-the-art methods demonstrate that our E4S outperforms existing methods in preserving texture, shape, and lighting. Our implementation is available at \url{https://github.com/e4s2024/E4S2024}.
\end{abstract}


\let\thefootnote\relax\footnotetext{\IEEEcompsocthanksitem $*$ Eual contribution, 
$\dagger$ Corresponding authors.}

\section{Introduction}
\label{sec:introduction}

\IEEEPARstart{F}{ace} swapping 
aims at not only transferring the identity information (e.g., shape and texture of facial components) of a source face to a given target face,  but also keeping the identity-irrelevant attribute information unchanged (e.g., expression, head pose, lighting, and background). It has broad masses of applications in movie composition, computer games, and virtual human broadcasting, attracting considerable attention in the field of computer vision and graphics.


There are two long-standing challenges in face swapping field. The first is \textbf{identity preservation}, \ie, making the swapped results faithfully preserve the facial characteristics of the source person. Most mainstream approaches
~\cite{chen2020simswap, li2019faceshifter, wang2021hififace, luo2022styleface, xu2022styleswap} extract the global identity-related features and inject them into the face generation process, where these identity features are obtained via a pre-trained 2D face recognition network~\cite{deng2019arcface}  or a 3D morphable face model (3DMM)~\cite{blanz19993dmm, deng2019accurate}. Nonetheless, their swapped results look like a third person, which resemble both the source and target faces.
Here, we argue the potential reason is current identity extractors are mainly designed for classification rather than generation~\cite{shiohara2023blendface}, thus cannot capture some informative and important facial visual details. Moreover, when a 3D face model takes one single image as input~\cite{li20233dswap}, it can hardly achieve precise shape recovery robustly.

The second challenge is how to deal with \textbf{facial occlusion.} 
It is common that some face regions are occluded by hair (or eyeglasses) in the input images. An ideal swapped result should maintain the hair (or eyeglasses) of the target. That is, it needs to generate the occluded facial regions for the source face. To deal with occlusion, FSGAN~\cite{nirkin2019fsgan} trains an inpainting sub-network to generate the missing pixels of the source. However, their inpainted faces are blurry. 
FaceShifter~\cite{li2019faceshifter} designs a refinement network to recover the occluded region in the target. Nonetheless, the refinement network tends to bring back certain identity information of the target to the source.

Although past efforts have achieved promising results, the above challenges still pose the main obstacles that prevent fine-grained face swapping.
Here comes to a question that can we build a face swapping framework to handle the aforementioned challenges at once? In this paper, we will explore such feasibility. We propose to rethink face swapping from a new perspective of fine-grained face editing, \ie, \textit{``editing for swapping'' (E4S)}. The key insight behind this is that \textit{given the disentangled shape and texture of individual facial components from two faces, face swapping can be transformed into the problem of local shape and texture replacement between them.}

Inspired by the previous fine-grained face editing approach~\cite{lee2020maskgan}, we apply separated component masks for local feature extraction. Then, we recompose local shape and texture features, which are fed into a mask-guided generator to synthesize the swapped result. Before conducting shape and texture swapping, we use a face reenactment model~\cite{wang2021faceVid2Vid} to drive the source person to have the same pose and expression as the target. The unique advantage of our E4S framework is that the facial occlusion challenge can be naturally addressed by facial masks since the face parsing network~\cite{yu2021bisenetv2} can provide labels of each face region. Our mask-guided generator can fill out the swapped masks with the swapped texture features adaptively. It does not rely on additional dedicated modules as in previous efforts~\cite{li2019faceshifter,nirkin2019fsgan}.

Given that StyleGAN~\cite{karras2020styleGAN2} has achieved striking performance on high-quality image editing, the core of our E4S framework takes advantage of a pre-trained StyleGAN~\cite{karras2020styleGAN2} for the disentanglement of shape and texture.
However, the existing GAN inversion methods~\cite{shen2020interfacegan, richardson2021psp, wang2022high} only performs global attribute editing (\eg, age, gender, and expression) in the global style space of StyleGAN, while the local control has not yet been explored. To achieve this goal, we propose a novel Regional GAN Inversion (RGI) method that incorporates facial masks into style embedding and introduces a regional $\mathcal{W}^{+}$ space, indicated as $\mathcal{W}^{r+}$. Specifically, we design a mask-guided multi-scale encoder that maps an input face into the style space of StyleGAN, where we make each facial component have a set of style codes for different layers of the generator. Besides, based on given facial masks, we propose a mask-guided injection module using style codes to manipulate the feature maps in the generator. In a word, we use style codes and masks to represent the texture and shape and conduct their disentanglement.



Besides shape and texture, we argue that lighting is also crucial for face swapping. If the lighting condition of the source and target is extremely different, directly transferring the source facial color space into the target image will produce an unnatural result. To deal with this, we split the lighting transferring problem into two steps. First, we inject the source skin texture along with its lighting into the target face to obtain a naive swapped face. Second, we transfer the target lighting to the swapped face by conducting face re-coloring under the guidance of the target color space. Concretely, we train a face re-coloring network in a self-supervised manner. It is trained by paired data with the same identity, where one is a grayscale face while the other is the corresponding colored face. This two-step scheme preserves the source skin details while yielding a natural lighting of the swapped images.


The remaining problem is how to obtain a source-consistent swapped face shape since it is difficult to recompose a perfect swapped mask during exchanging masks, especially when the face shape of source and target have a large gap. To tackle this, we propose a face inpainting network to refine the face shape, which is trained through an unsupervised manner. We synthesize the paired data by erasing random areas from the original faces and train the network to predict the erased parts through visible ones based on a mismatch mask. Additionally, based on the modulation convolution~\cite{viazovetskyi2020stylegan2}, we propose to use the area ratio of mismatch regions to adaptively control the generation process. During inference, guided by those mismatched mask regions, the inpainting network can refine the face shape to make it preserve the shape of the source face. 
In summary, the main contributions of this paper are:
\begin{Itemize}

\item We propose E4S, a fine-grained face swapping framework from a new perspective of face editing. Our high-fidelity face swapping can preserve source identity and 
target lighting and dealing with  occlusion challenge.  \footnote{The implementation is at \href{https://github.com/e4s2024/E4S2024}{https://github.com/e4s2024/E4S2024}. The project page is available at \href{https://e4s2024.github.io/}{https://e4s2024.github.io/}.}

\item We propose a Regional GAN Inversion (RGI) method for the explicit disentanglement of shape and texture based on a pre-trained StyleGAN.

\item We solve the lighting transferring problem by splitting it into two parts. In the first part, we integrate the source skin along with the source lighting into the target image. In the second part, we train a face re-coloring network in an unsupervised manner, which re-colors the swapped image guided by the target color space.

\item We additionally designed a face inpainting network modulated by the mismatch region ratio, which keeps the swapped face shape consistent with the source one.

\end{Itemize}

\section{Related Work}
\noindent{\textbf{GAN Inversion.}}
GAN Inversion aims to map an image to its corresponding latent code that can faithfully reconstruct the input, which is useful for image editing since the inverted code can be modified and then fed into the generator to produce the desired output. Existing approaches for StyleGAN inversion can be broadly categorized into three groups: learning-based~\cite{richardson2021psp, tov2021e4e, alaluf2021restyle, wang2022high, yao2022FSspace, yao2021latent}, optimization-based~\cite{abdal2019image2stylegan, abdal2020image2stylegan++, kang2021GANforOORimages, saha2021loho, zhu2021barbershop}, and hybrid methods~\cite{zhu2020indomainGAN}. Learning-based methods involve training an encoder to map the image to the latent space, whereas optimization-based methods directly optimize the latent code to minimize the reconstruction error. Although optimization-based methods tend to yield better results, they are more computationally expensive than learning-based methods. Hybrid methods seek to balance these two approaches by using the inverted code as a starting point for further optimization.

Existing GAN inversion methods always perform global editing, allowing for modifications such as changes in pose, gender, and age. However, they cannot precisely control local facial features. To address this limitation, we utilize Regional GAN Inversion, which employs a novel $\mathcal{W}^{r+}$ latent space based on a pre-trained StyleGAN. This new approach enables high-fidelity local editing of facial components, filling an important gap in existing methods.

\noindent{\textbf{Face Swapping.}}
Face-swapping approaches can be broadly categorized into two classes: source-oriented and target-oriented~\cite{chen2020simswap}. Source-oriented methods~\cite{blanz2004exchangingFace,bitouk2008faceSwapping,nirkin2018OnFaceSeg,nirkin2019fsgan,nirkin2022fsganv2} initiate from the source, aiming to transfer the target's attributes to it. Early techniques in this category can be traced back to~\cite{blanz2004exchangingFace}, which estimated 3D shape and relevant scene parameters for pose and lighting alignment. Later, \cite{nirkin2018OnFaceSeg} claimed that 3D shape estimation was unnecessary and applied face segmentation for face swapping. Recently, FSGAN~\cite{nirkin2022fsganv2,nirkin2019fsgan} employed a two-stage pipeline where a reenactment and inpainting network addressed pose alignment and occlusion issues, respectively.
As contrast, target-oriented methods~\cite{korshunova2017fastFaceSwap,bao2018IPGAN,chen2020simswap,li2019faceshifter,wang2021hififace,xu2022region,kim2022smoothswap, ren2023wscswap,zhao2023diffswap} start with the target, intending to import the source identity. Generally, these approaches maintain the source identity using a pre-trained face recognition model~\cite{shiohara2023blendface} or 3DMMs~\cite{li20233dswap}. However, as the recognition model is trained for classification and 3DMMs lack accuracy and robustness, these methods fail to fully capture identity-related details for generation.

For StyleGAN-based face swapping, \cite{zhu2021MegaFS, luo2022styleface, xu2022styleswap, jiang2023styleipsb} leverages prior knowledge from the pre-trained StyleGAN, increasing image resolution to $1024^2$. StyleFusion~\cite{kafri2021stylefusion} performs latent fusion within the $\mathcal{S}$ space~\cite{collins2020editingInStyle,chong2021retrieveInStyle}, allowing for controllable generation of local semantic regions. Nonetheless, the shape and texture of each facial region remain entangled in the $\mathcal{S}$ space. Besides, \cite{xu2022region} proposes a region-aware projector for adaptively transferring source identity to the target face. HiRes\cite{xu2022high} utilizes an additional encoder-decoder for multi-scale target feature aggregation. However, these two techniques do not support fine-grained and selective swapping.

Our \textit{E4S} belongs to the source-oriented group. Taking inspiration from mask-guided face editing~\cite{park2019semantic,lee2020maskgan,zhu2020sean,chen2021sofgan}, we reframe face swapping as editing shape and texture for all facial components. We propose to explicitly disentangle facial components' shape and texture using the RGI method for better identity preservation, instead of relying on a face recognition model or 3DMMs.
    \begin{figure*}[t]
      \centering
    \includegraphics[width=1.0\linewidth]{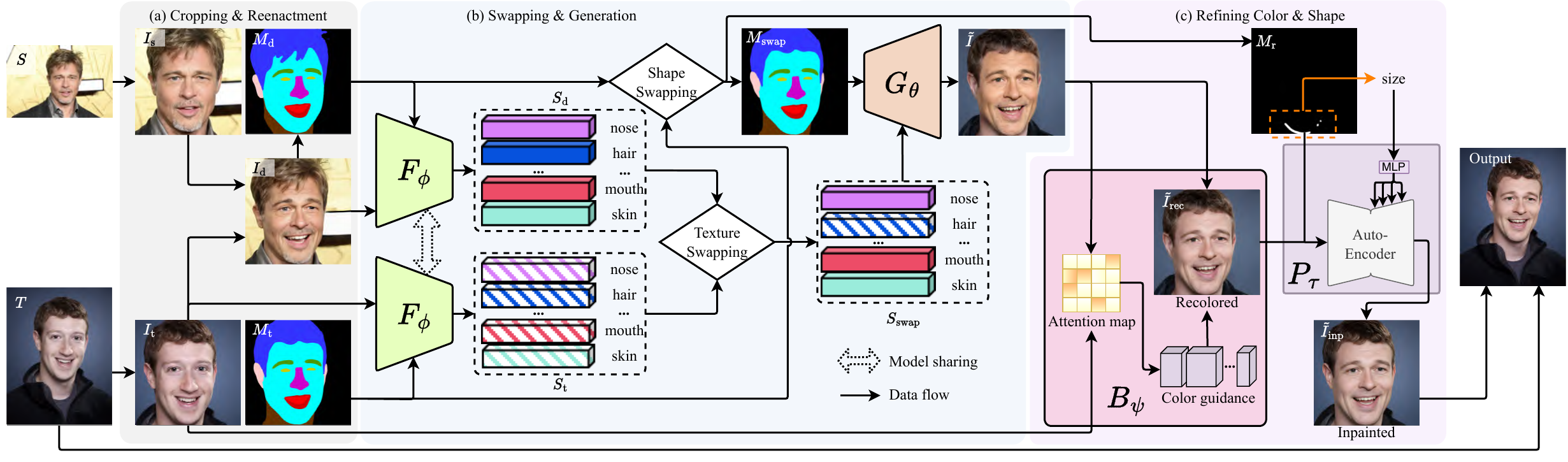}
       \caption{E4S framework overview. (a) We first crop the face region of the source $S$ and the target $T$ to obtain $I_{\rm{s}}$ and $I_{\rm{t}}$. Then, a reenactment network $G_{\rm{r}}$ encourages $I_{\rm{s}}$ to have a similar pose and expression towards $I_{\rm{t}}$,  obtaining the driven image $I_{\rm{d}}$. The segmentation masks of $I_{\rm{t}}$ and $I_{\rm{d}}$ are also estimated. (b) Then, the driven and target pairs $(I_{\rm{d}}, M_{\rm{d}})$ and $(I_{\rm{t}}, M_{\rm{t}})$ are fed into the mask-guided encoder $F_{\phi}$ to extract the per-region style codes to depict the texture respectively, producing texture codes $S_{\rm{d}}$ and $S_{\rm{t}}$. Next, we exchange the masks and the corresponding texture codes, obtaining $S_{\rm{swap}}$ which is then sent to the pre-trained StyleGAN generator $G_{\theta}$ with a mask-guided injection module to synthesize the naive swapped face $\tilde{I}$. (c) Finally, we propose a refinement stage, which includes a face re-coloring network $B_{\psi}$ for transfering the target lighting to $\tilde{I}$, and a face inpainting network $P_{\tau}$ for preserving a consistent shape with source face. }
       \label{fig:pipeline}
       \vspace{-0.2cm}
    \end{figure*}

\section{Method}
In this section, we first introduce our proposed editing-for-swapping (E4S) framework and elaborate on each inside step in Sec.~\ref{sec:swappingFace}. Then, we explain the proposed region GAN inversion (RGI) method for the disentanglement of the shape and texture of facial components in Sec.~\ref{sec:regionalGANInversion}.  
Each core module of the RGI is detailed subsequently. 
Next, we introduce the face recoloring network $B_{\psi}$ for maintaining the target skin tone and lighting, and the face inpainting network $P_{\tau}$ for the unnatural results caused by the possible shape mismatch during mask exchange in Sec.~\ref{sec:enhence}.
After that, we present the loss functions for training in Sec.~\ref{sec:loss}. 

\subsection{Editing For Swapping (E4S) framework}
\label{sec:swappingFace}
The pipeline of our face swapping framework is illustrated in Fig. \ref{fig:pipeline}, which mainly consists of three phases inside: (a) cropping and reenactment, (b) swapping and generation, and (c) lighting transfer and face shape inpainting.

\subsubsection{\textbf{Cropping and Reenactment}}
We first use the dlib toolbox~\cite{dlib09} to crop the face region of the source image $S$ and target image $T$ respectively, obtaining the cropped faces $I_{\rm{s}}$ and $I_{\rm{t}}$. 
Then, we follow the previous method~\cite{karras2019styleGAN} to align the cropped face and resize it into 1024$\times$1024 resolution.

To drive $I_{\rm{s}}$ to obtain the similar pose and expression as $I_{\rm{t}}$, we employ a pre-trained face reenactment model FaceVid2Vid~\cite{wang2021faceVid2Vid}, obtaining a driven face $I_{\rm{d}}$. 
The face reenactment processing can be expressed as:
\begin{equation}
    I_{\rm{d}} = G_r(I_{\rm{s}}, I_{\rm{t}}),
\end{equation}
where $G_r$ is the FaceVid2Vid model. 
Then, the segmentation masks $M_{\rm{t}}$ of the target face $I_{\rm{t}}$ and $ M_{\rm{d}}$ of the driven face $I_{\rm{d}}$ are estimated by an off-the-shell face parser~\cite{zllrunning2013faceParser}.
In this way, the target and driven pairs ($I_{\rm{t}}, M_{\rm{t}}$) and ($I_{\rm{d}}, M_{\rm{d}}$) can be formed, where each segmentation mask belongs to one of the 19 semantic categories. 
For brevity, we aggregate the categories of symmetric facial components, bringing 12 categories totally, \ie, \textit{background, eyebrows, eyes, nose, mouth, lips, face skin, neck, hair, ears, eyeglass, and ear rings}.


\subsubsection{\textbf{Swapping and Generation}}
\label{subsec:faceSwapping}
After the cropping and reenactment process, we are ready to fulfil the face swapping process. We first feed the driven pair ($I_{\rm{d}}, M_{\rm{d}}$) and the target pair ($I_{\rm{t}}, M_{\rm{t}}$) into a mask-guided multi-scale encoder $F_{\phi}$ respectively, obtaining the style codes to represent the texture of each facial region. 
This step can be described as:
\begin{equation}
    S_{\rm{t}} = F_{\phi}(I_{\rm{t}}, M_{\rm{t}}),\quad S_{\rm{d}} = F_{\phi}(I_{\rm{d}}, M_{\rm{d}}),
\end{equation}
where $S_{\rm{t}}$ and $S_{\rm{d}}$ denote the extracted texture codes of the target and driven face, respectively. 
The detailed modules of the encoder $F_{\phi}$ are introduced in Sec.~\ref{sec:regionalGANInversion}. 
Then, we exchange the texture codes of several facial components of $S_{\rm{t}}$ with those of $S_{\rm{d}}$, resulting in the recomposed texture codes $S_{\rm{swap}}$. 
The components that must be exchanged from $S_{\rm{d}}$ are: \textit{eyebrows, eyes, nose, mouth, lips, face, skin, neck, and ears}.
\CUT{
One may be confused about the exchange of \textit{skin}, 
since this paper aims to maintain the skin tone and lighting of the target.
The reason we use the \textit{skin} component from the source is replacing the texture codes of skin from the source face with the target one can not bring satisfactory results. Thus, we turn to a skin recoloring network as refinement, rather than keeping the skin color of the target directly in the swapping process.
}
Note that due to the entanglement of skin tone and lighting, we split the skin color processing into two steps: (i) transferring the source skin texture including skin tone and lighting from the source to the swapped faces (Sec.~\ref{sec:regionalGANInversion}); (ii) recoloring the swapped results under the guidance of the target color space (Sec.~\ref{sec:enhence}). The experimental results in Fig.~\ref{fig:ablation_recolor} show that our scheme outperforms previous methods, especially on the consistency of source identity.


To achieve face swapping, besides of texture swapping, shape swapping is also needed. Considering face shape can be indicated by facial masks, we start with an empty mask $M_{\text{swap}}$ as a canvas and then complete the mask recomposition.
Specifically, we first keep the neck and the background from the target mask $M_{\rm{t}}$ and stitch their masks onto $M_{\text{swap}}$. 
Next, we stitch the inner face regions of the driven mask $M_{\rm{d}}$, including \textit{face skin, eyebrows, eyes, nose, lips, and mouth}. 
Finally, we stitch the \textit{hair, eye glasses, ear, and ear rings} from the target mask $M_{\rm{t}}$ to the $M_{\text{swap}}$.


Next, we feed the recomposed mask $M_{\text{swap}}$ and texture codes ${S_{\text{swap}}}$ into the StyleGAN generator $G_{\theta}$ with a mask-guided style injection module to generate the naive swapped face $\tilde{I}$:
\begin{equation}
    \tilde{I} = G_{\theta}(M_{\text{swap}}, S_{\text{swap}}).
\end{equation}
The details of the generator $G_{\theta}$ will be introduced in the Sec.~\ref{sec:regionalGANInversion}. Note that our E4S does not need to train an extra sub-network to deal with the occlusion and can naturally achieve more accurate occlusion recovery than the existing methods FSGAN~\cite{nirkin2019fsgan} and FaceShifter~\cite{li2019faceshifter}. This is because the generator $G_{\theta}$ can fill out the occlusion pixels with the swapped texture features adaptively according to masks.

\subsubsection{\textbf{Lighting Transfer and Face Shape Inpainting}}
Since the disentanglement of texture and shape solely cannot guarantee a lighting-natural swapped result.
That is, the lighting of the source contained in the texture information is hard to be disentangled well~\cite{zhu2020aot}, leading to the lighting leakage to the swapped faces.
To tackle this issue and harmonize the lighting results, on the basis of the aforementioned transferring skin texture, 
we additionally propose a face-recoloring network $B_{\psi}$ to transfer lighting from the target face $I_{\rm{t}}$ to the naive swapped face $\tilde{I}$, which is elaborated in Sec.~\ref{sec:re-coloring}. 
\CUT{
Then, the swapped face $\tilde{I}$ and target image $T$ are blended together via Multi-Band Blending~\cite{mb-blending,kim2018image}, producing \lmm{the naive swapped image $F_{swap}$. (Is it right?)} 
}
Then, considering that the potential mismatch during the mask recombination process of $M_{\rm{d}}$ and $M_{\rm{t}}$ would bring difficulties for the shape preservation of source face, we propose an inpainting network $P_{\tau}$, which operates on pixel level and maintain the source face shape by the guide of mismatched region masks. we give the detailed description of the inpainting network $P_{\tau}$ in the Sec.~\ref{sec:inpainting}.


\subsection{Disentangling Shape and Texture}
\label{sec:regionalGANInversion}
The core of the proposed E4S framework is disentangling the per-region texture and its corresponding shapes. To pursue a better disentanglement of shape and texture as well as high-resolution and high-fidelity generation, we resort to the powerful generative model StyleGAN that can generate images with 1024$\times$1024 resolution naturally. Specifically, we develop a GAN inversion method rather than  training StyleGAN from scratch, which also avoids training instability.

\begin{figure}[t]
  \centering
  \includegraphics[width=1.0\linewidth]{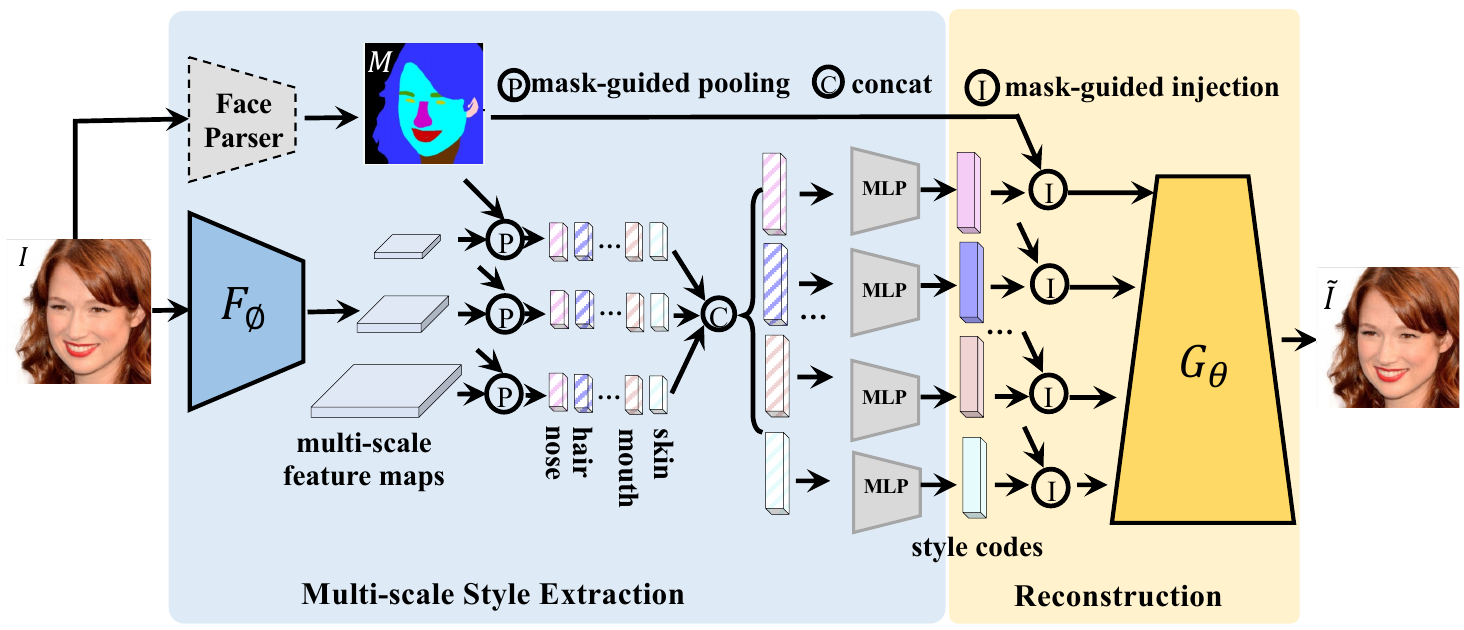}
  \caption{Overview of the proposed RGI. The input face $I$ together with the corresponding segmentation map $M$ are fed into a multi-scale encoder $F_{\phi}$ to extract the per-region texture vectors. The multi-scale texture vectors are then concatenated and passed through some MLPs to get the style codes resident in the latent space of StyleGAN. The regional style codes and the mask $M$ are used by our mask-guided StyleGAN generator to produce the reconstructed face $\tilde{I}$.}
  \label{fig:RGI}
\end{figure}

Although there are quite a few GAN inversion methods ~\cite{richardson2021psp, tov2021e4e, yao2022FSspace} have been proposed for face editing in the style space, they constantly focus on global facial attribute editing (\textit{eg}, age, pose, expression, etc.) and do not pay attention to disentanglement of shape and texture for local editing. To fill this gap, we propose a novel Regional GAN Inversion (RGI) method for such a disentanglement, which incorporates facial masks into the style embedding and the generation process. The overview of RGI is in Fig.~\ref{fig:RGI}.

\subsubsection{Mask-guided Style Extraction}
\label{sec:encodingStyle}
Given an image $I$ and its segmentation mask $M$, we first leverage a multi-scale encoder $F_{\phi}$ to take the image $I$ as input and produce feature maps at different levels:
\begin{equation}
[F_1,F_2,...,F_N] = F_{\phi}(I),
\end{equation}
where N represents the number of scales and
$F_{\phi}$ indicates a convolution network with multiple layers. Then, the multi-scale features for each individual facial region can be obtained via the feature maps $[F_1,F_2,...,F_N]$ and the mask $M$.
Concretely, we downsize the mask $M$ to the same scale with each feature map $F_i$, and then use the average pooling operation on $F_i$ to aggregate features for each facial region:
\begin{equation}\label{eq:avgpooling}
v_{ij} = \text{Average}(F_i \odot (\lfloor M \rfloor_i == j)), \forall j \in \{1,2,...,C\},
\end{equation}
where $C$ denotes the number of segmentation categories, $\odot$ represents the Hadamard product, and $\lfloor M \rfloor_{i}$ indicates the downsized mask with the same size as $F_i$. 
Further, we concatenate the multi-scale feature vectors $\{v_{ij}\}_{i=1}^{N}$ of the region $j$ and put them into an MLP, thus obtaining the style codes, which can be described as:
\begin{equation}
    s_{j} = \text{MLP}([v_{1j};v_{2j};...;v_{Nj}]),
\end{equation}
where $s_j$ indicates the style codes of the $j$-th facial region. 
Then, the generator $G_{\theta}$ takes as input the style codes and the mask $M$ are fed into the generator to produce the navie swapping face $\tilde{I}$.
Formally, we define $s\in\mathbb{R}^{C \times 18 \times 512}$ as the proposed $\mathcal{W}^{r+}$ space.






\subsubsection{Mask-guided Style Injection}
\label{sec:localstylegan}
\begin{figure}[t]
  \centering
  \includegraphics[width=\linewidth]{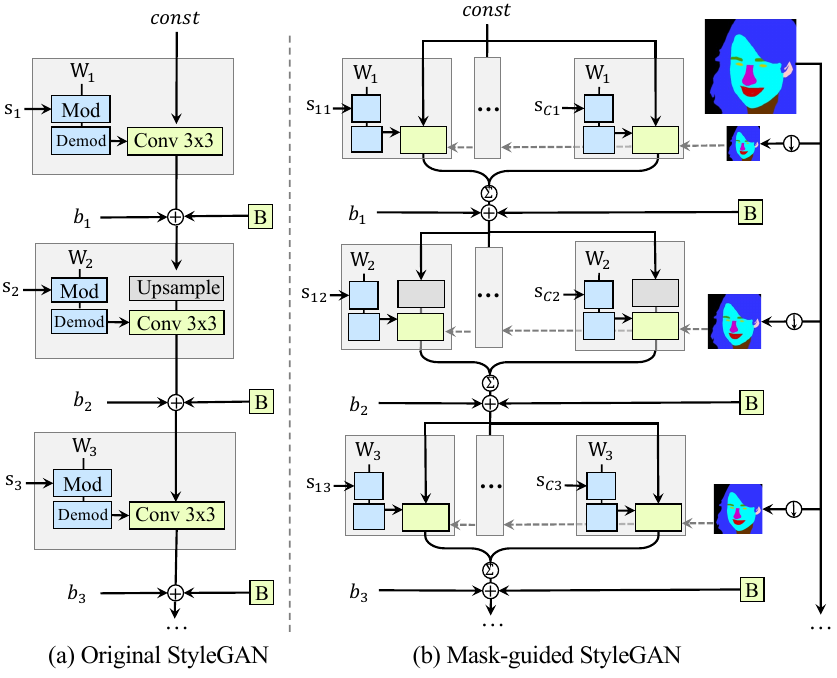}
  \caption{The comparison of the original StyleGAN and the proposed mask-guided StyleGAN. (a) The original StyleGAN contains consecutive convolution blocks. Each block contains a modulation, a demodulation, and a convolution layer. $W$ and $b$ denote the learnable kernel weights for each block, and $s$ denotes the style code. \fbox{B} is noise broadcast operation. An upsampling layer is used between every two blocks. (b) Our mask-guided StyleGAN regionally extends the convolution block. We sum up the intermediate feature maps of each region using its segmentation mask which is downsized in advance. }
\label{fig:mask_stylegan}
  \end{figure}

The original StyleGAN contains a style-based generator taking 18 style codes with the dimension of 512 as input. The style codes are used to manipulate the feature maps of the 18 intermediate layers. As shown in Fig. \ref{fig:mask_stylegan} (a), the generator, consisting of a serial of style blocks, takes a constant feature map with the spatial size of 4$\times$4 as input. Each style block includes a modulation, a demodulation, 3$\times$3 convolution layer, and a noise layer \fbox{B} to increase diversity. Here, the learnable kernel weights and bias in each block are denoted as $W$ and $b$. $W$ would be scaled by its corresponding style code $s$ with the shape of 1$\times$512 before the convolution layer. An additional upsampling layer by the factor of two is adopted between every two style blocks, increasing the feature resolution.

In this paper, we aim to extract regional style code that controls the local appearance of the corresponding face component precisely along with its mask, which is opposite to the way of style code in the original StyleGAN, which globally controls the appearance of the output image.
To achieve this, we overhaul the style block of the original StyleGAN to a mask-guided style block, which is based on a given mask. Fig. \ref{fig:mask_stylegan} (b) illustrates
the schematic operations of our proposed mask-guided style injection.
Specifically, we sum up the intermediate feature maps with the guidance of per-region mask as:
\begin{gather}
F_{l} =  \sum_{j=1}^{C}(F_{l-1} * W'_{jl}) \odot (\lfloor M \rfloor_{l} == j),  \forall\  l \in \{1,2,...,K\}, \\
W'_{jl} = Demod(Mod(W_l, s_{jl})) \label{eqn:modulation},
\end{gather}
where $F_{l-1}$ and $F_{l}$ indicate the input and output feature maps of $l$-th layer.
$W'_{jl}$ represents the scaled kernel weights for the $j$-th component in the $l$-th layer, and $*$ denotes the convolution operation. Similar to Eq. \ref{eq:avgpooling}, the $\lfloor M \rfloor_{l}$ is the downsized mask corresponding to the $l$-th layer.
Following the same modulation and demodulation as the original StyleGAN, we extend the style modulation regionally. In Eq.~\ref{eqn:modulation}, $W_{l}$ denotes the original kernel weights for the $l$-th layer, and the $s_{jl}$ indicates the style code of $j$-th component for the $l$-th layer.

It is worth noticing that we only inject the mask into the first $K$ layers of the StyleGAN and do not use the mask-guided style block for the last ($18-K$) layers. 
The reason here is twofold: (i) we experimentally find that the reconstructed images have few visual differences when $K$ is greater than 13; 
(ii) since the resolution of the last ($18-K$) layers is large (\ie, $512^2-1024^2$), the training cost would be lowered when mask-guided style injection is not conducted in these layers.
Thus, we set $K=13$ as the default in all the experiments.




\begin{figure}[t]
  \centering
\includegraphics[width=0.95\linewidth]{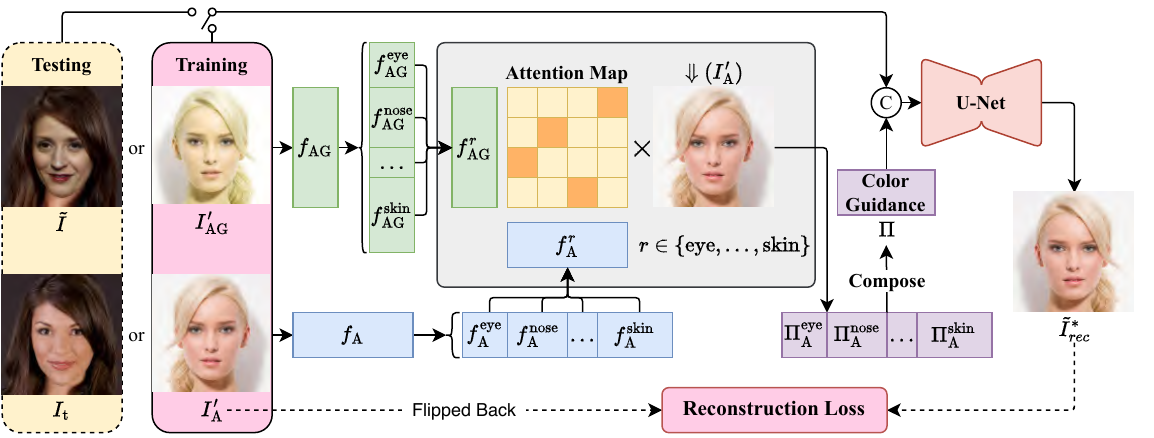}
  \captionsetup{belowskip=-3pt}
  \caption{Illustration of our re-coloring network $B_{\psi}$. During training, we first obtain the random re-colored $I'_{\rm{AG}}$ and flipped $I'_{\rm{A}}$ and extract their features $f_{\rm{AG}}$ and $f_{\rm{A}}$ by FPN~\cite{zhang2020cross}. For each region $r$, we calculate an attention map between $f_{\rm{AG}}$ and $f_{\rm{A}}$, which is then multiplied with the downsampled $I'_{A}$ and composed to generate color guidance $\Pi$. Finally, a U-Net takes as input the concatenated $\Pi$, the input images $I'_{\rm{AG}}$, $I'_{\rm{A}}$ and their segmentation masks, generating the re-colored result $\tilde{I}^*_{\rm{rec}}$, which is used to calculate the reconstruction loss with the flipped $I'_{\rm{A}}$. During testing, we adopt a same scheme to transfer the color from the target $I_{\rm{t}}$ to the naive swapped face $\Tilde{I}$, resulting in the re-colored swapped result $\tilde{I}^*_{\rm{rec}}$.
  }
  \label{fig:re_coloring}
\end{figure}

\subsection{Lighting Transfer and Shape Inpainting}
\label{sec:enhence}
Although our E4S framework provides a unique face swapping insight (i.e, disentangling shape and texture), there are still some situations that it cannot handle well.
{(i)} When the lighting difference between source and target is huge, reconstructing the face of source in the ambient lighting of target would produce disharmonious results.
{(ii)} During recomposing the mask $M_{\text{swap}}$, there would inevitably be some unmatched mask regions. For instance, for those positions belong to the inner face of the target, but not to that of the source, to maintain the shape of the source, these unmatched face regions should be injected with the style code of non-inner face category (hair, neck, etc.). However, since different unmatched regions may correspond to different injected style codes, it is difficult to arrange them in advance.
Furthermore, even if the mismatch regions are injected with the correct style code and yield a perfect recomposed mask $M_{\text{swap}}$, guided by this mask, the content generated in the mismatch regions tends to conflict with
the original target image $T$. It brings difficulty on blending the naive swapped face
$\tilde{I}$ and target $T$.

To address the above issues, we propose two refinement techniques for more fine-grained and lighting-consistent face swapping, which consist of:
transferring lighting condition and skin tones from the target to the swapped result with an additional face re-colorer $B_{\psi}$, and training an additional inpainting network $P_{\tau}$ to restore mismatch-mask regions.

\subsubsection{\textbf{Face Re-coloring}}
\label{sec:re-coloring}
Preserving the target lighting condition is challenging, especially when the illumination of the source is leaked into the result. Disentangling the illumination requires recovering material, geometry, and lighting of a scene, making this issue ill-posed~\cite{zhu2020aot}.
To deal with the re-lighting challenge, we reformulate the re-lighting problem as a re-coloring problem and propose to transfer the color from the target to the swapped face by our re-colorer $B_{\psi}$.
Although the swapped face color should be determined by both the source skin tone and target environment lighting, we simplify the re-lighting problem by transferring the target color to the swapped face.
In practice, we found this simplification leads to reasonable and admirable results, improving the result fidelity, as shown in Fig.~\ref{fig:ablation_recolor}.

However, it is impossible to collect annotated paired data for the training of transferring the color of a reference face to the objective face.
It may take great expense and huge labor to manually paint reasonable output. 
Thus, we turn to perform self-supervised training, where we set the face to be colored and the reference face as the same identity during training.
Specifically, for a given RGB face $I_{\rm{A}}$, we convert it into a grayscale image $I_{\rm{AG}}$.
As seen in Fig.~\ref{fig:re_coloring}, we first conduct random color augmentation ${\rm{CA}}$ on the grayscale image $I_{\rm{AG}}$ and a random horizontal flip  augmentation ${\rm{FA}}$ on the original image $I_{\rm{A}}$ respectively:
\begin{equation}
\begin{aligned}
{I_{\rm{AG}}'}={\rm{CA}}(I_{\rm{AG}}), {I_{\rm{A}}'}={\rm{FA}}(I_{\rm{A}}), 
\end{aligned}
\end{equation}
The reason for using additional flip augmentation is to prevent the network from copying the pixels from the same position directly.
Then, we obtain feature representation of $I_{\rm{AG}}'$ and $I_{\rm{A}}'$ using feature pyramid network FPN~\cite{zhang2020cross}:
\begin{equation}
\begin{aligned}
f_{\rm{AG}}={\rm{FPN}}({I_{\rm{AG}}'})\in\mathbb{R}^{N\times C}, 
f_{\rm{A}}={\rm{FPN}}(I_{\rm{A}}')\in\mathbb{R}^{N\times C},
\end{aligned}
\end{equation}
where $C$ indicates the number of channels and $N$ is the spatial zone of features.
The next step is to compute the correlations between extracted features in each spatial location. Specifically, we calculate the correlations among the same semantic region in facial mask $M_{\rm{t}}$, avoiding correlation calculation among those uncorrelated regions. The process can be expressed as:
\begin{equation}
S^r= f_{\rm{AG}}^r {f^r_A}^{\top} \in \mathbb{R}^{N^r \times N^r}.
\end{equation}
where $f_{\rm{AG}}^r$ and $f_{\rm{A}}^r$ are the corresponding feature of each region $r \in$ \{
\textit{eyebrows, eyes, nose, mouth, lips, face, and skin}\}.
$S^r$ indicates pair-wise correlation of each facial region, and $S^r_{i,j}$ is similarity between $i$-th pixel in $f_{\rm{AG}}^r$ and $j$-th pixel in  $f_{\rm{A}}^r$.
$N_r$ denotes the number of pixels of each region. 
Next, the correlation matrix $S^r$ is normalized by softmax layer and then multiplied with the pixels of downsampled $I_{\rm{A}}'$ in region $r$, which can be expressed as:
\begin{equation}
\label{softmax}
\begin{aligned}
\Pi^r = \frac{{S^r}}{\sum_{j=1}^{N^r}{S^r_{.,j}}} ({\Downarrow}(I_{\rm{A}}')^r) \in\mathbb{R}^{N^r\times 3},
\end{aligned}
\end{equation}
where $\Pi^r$ denotes the color guidance in region $r$, and ${\Downarrow}(\cdot)$ indicates the downsampling operation.
Combining all the $\Pi^r$, we can obtain the full facial color guidance $\Pi \in \mathbb{R}^{N\times 3}$.
Then, we concatenate the representation $I_{\rm{AG}}'$, $I_{\rm{A}}'$, upsampled color guidance ${\Uparrow}(\Pi)$, and the semantic segmentation mask $M_A$ (ommited in Fig.~\ref{fig:re_coloring}) of $I_{\rm{A}}$, and pass them to train a recoloring UNet, where ${\Uparrow}(\cdot)$ indicates the upsampling operation.
Our goal here is to distil color from the colored image $I_{\rm{A}}'$ to $I_{\rm{AG}}'$.
Once this re-colorer $B_{\psi}$ is trained, we can transfer the {color} of the target $I_{\rm{t}}$ to $\tilde{I}$ as: 
\begin{equation}
\tilde{I}^*_{\rm{rec}} = B_{\psi}(\tilde{I}, M_{\rm{d}}, I_{\rm{t}}, M_{\rm{t}}),
\end{equation}
where $M_{\rm{d}}$ and $M_{\rm{t}}$ are semantic segmentation masks of the swapped face $\tilde{I}$ and target $I_{\rm{t}}$, respectively.

Furthermore, since the high computational cost of calculating cosine similarity in 1024$\times$1024 resolution (e.g. calculating cosine similarity matrix between the \textit{skin} parts of two 1024$\times$1024 faces requires 64GB memory in average), we train the U-Net in 256$\times$256 resolution. 
Although the memory is more efficient, the produced $256$ re-coloring result $\tilde{I}^*_{\rm{rec}}$ looks very blur when we directly resize it to 1024$\times$1024 resolution.
Therefore, we use a face super-resolution method~\cite{zhou2022codeformer} to enhance the low-resolution $\tilde{I}^*_{\rm{rec}}$ and paste the enhanced recolored image to $\tilde{I}$ based on a low-pass mask $M_{\rm{rec}}$ where the high-frequency pixels is removed by a Sobel filter,
where the final enhanced and pasted recolored output is denoted as $\tilde{I}_{\rm{rec}}$, as shown in Fig.~\ref{fig:ablation_recolor}.


\begin{figure}[t]
  \centering
  \includegraphics[width=\linewidth]{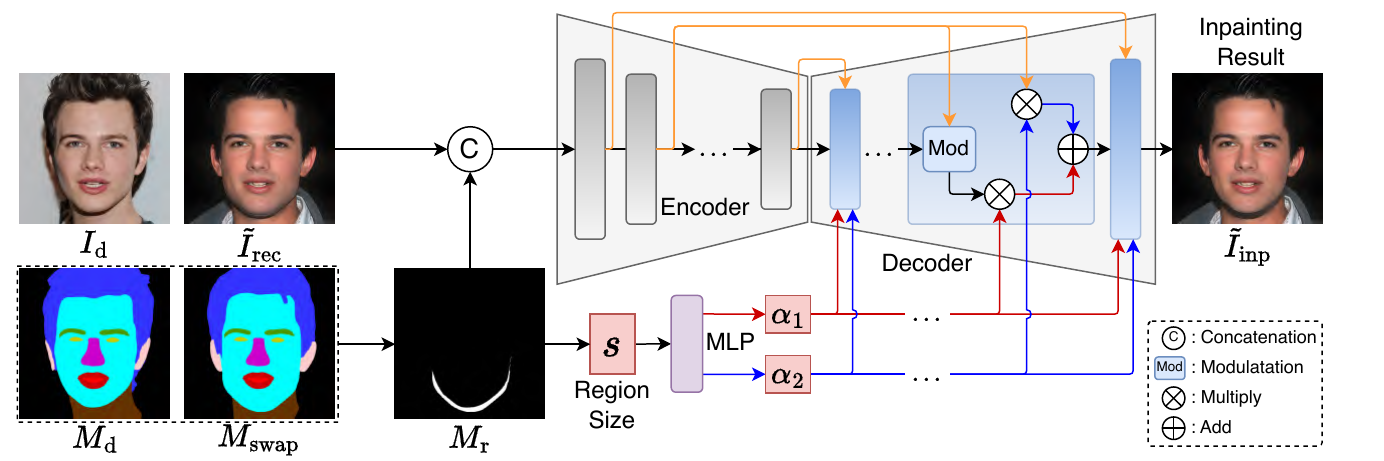}
  \caption{Illustration of inpainting network $P_{\tau}$, which adaptively inpaints the mismatch regions on the pixel levels. Given a driven face $I_{\rm{d}}$ and the pasted $\Tilde{I}_{\rm{rec}}$, we first calculate the mismatch regions mask $M_{\rm{r}}$. 
  Then the image $\Tilde{I}_{\rm{rec}}$ along with the mismatch mask $M_{\rm{r}}$ are fed to an auto-encoder to inpaint the mismatch pixels. The generation process of the decoder is modulated by the scale factor $\alpha_1$ and $\alpha_2$ extracted from the area ratio $s$ of mismatch regions. The final inpainting result is denoted as $\tilde{I}_{\rm{inp}}$. For simplicity, the normalization layers are omitted in this figure.}
  \label{fig:inpainting}
  \vspace{-2mm}
\end{figure}

\subsubsection{\textbf{Inpainting For Mismatch-mask Region}}
\label{sec:inpainting}
There would inevitably be quite a few mismatched regions during facial mask exchange between the driven face $M_{\rm{d}}$ and target $M_{\rm{t}}$ (Sec.~\ref{sec:swappingFace}).
Here, we divide facial mask into two categories: inner face region (\textit{eyebrows, eyes, nose, mouth, lips, face, and skin}) and non-inner face region (\textit{neck, ears, hair, earrings, and background}).
Then, there are two types of face mismatched pixels in the recomposed mask $M_{\text{swap}}$: 1) the positions belong to the inner face of the source, but not to that of the target; 
2) the positions belong to the inner face of the target, but not to that of the source.
In the first case, 
to keep face shape of the source, we directly use face mask of the source to fill those mismatched positions.
For the second case, we need to fill the position with a proper non-inner face category. In this paper, we  define a mismatched region $M_{\rm{r}}$ as the second case since it cannot be solved directly:
\begin{equation}
M_{\rm{r}}=\{(x, y) | M_{\rm{d}}^{\rm{inner}}[x, y] = 0 \ \&\ M_{\rm{t}}^{\rm{inner}}[x, y] = 1\},
\label{eq_hole}
\end{equation}
where $M_{\rm{d}}^{\rm{inner}}$ indicates the inner face region of driven face mask,
$M_{\rm{t}}^{\rm{inner}}$ denotes the inner face region of target mask.

A natural idea to solve the mismatched region is to design an algorithm to adaptively arrange each position in the region with a proper non-inner face category (e.g., nearest neighbour classifier). However, we empirically notice that
even if all mismatched regions are filled with the correct category, guided by this perfect recomposed mask, the content generated in this region tends to conflict with the original target image $T$. 
This is because some background information should be generated in those locations which once belonged to the face area.
Such conflict would lead to disharmony at the borders of the mismatched region after Multi-Band Blending.
Thus, this paper turns to train an inpainting network $P_{\tau}$ that fills the mismatched region in the blended result $\tilde{I}_{\rm{rec}}$ in the pixel-level rather than the mask-level.


The framework of our face-inpainting network $P_{\tau}$ is shown in Fig.~\ref{fig:inpainting}.
We train $P_{\tau}$ in an unsupervised manner, where we randomly erase the pixels around the face contour and then encourage the network to predict the erased regions. Specifically, given a face image $I$, we generate random mismatch masks and edit its skin belonging to these mismatch regions based on the editing ability of our RGI method, obtaining an edited face $I_{\rm{edit}}$.
Then, the inpainting network $P_{\tau}$ is trained by using $I_{\rm{edit}}$ as input and $I$ as the ground-truth supervision. 
During inference, given the recolored $\tilde{I}_{\rm{rec}}$ and the driven source $I_{\rm{d}}$, the mismatch regions $M_{\rm{r}}$ of these two faces is calculated by Eqn.~\ref{eq_hole}.
Then the inpainting network $P_{\tau}$ recovers the erased content, producing a refined result sharing the consistent face shape with the source.
Then the area ratio (ranging in $[0,1]$) of mismatch regions (white parts of $M_{\rm{r}}$ in Fig.~\ref{fig:inpainting}) is fed to an MLP, extracting the scale factor $\alpha_1$ and $\alpha_2$ which guide the generation process of the decoder.
Specifically, we use $\alpha_1$ and $\alpha_2$ to control the linear combination of the features extracted by the encoder and the corresponding modulated ones:
\begin{equation}
    x_{\rm{out}} = \alpha_1 {\rm{Mod}}(x_{\rm{in}}, x_{\rm{skip}}) + \alpha_2 x_{\rm{skip}},
\end{equation}
where the modulation process $\rm{Mod}(\cdot)$ follows the modulating convolution of StyleGAN~\cite{karras2020styleGAN2}, $x_{\rm{in}}$ and $x_{\rm{out}}$ denote the input and output feature of a decode layer, $x_{\rm{skip}}$ indicates the skip connection features from the corresponding layer, all the subscripts of layer index are omitted for convenience sake.
We only copy the inpainted pixels lying in mismatched regions to the input.
Consequently, the network $P_{\tau}$ is capable of
generating the inpainting results $\tilde{I}_{\rm{inp}}$ which preserves
the facial shape of the source without additionally seeking a perfect recomposed mask $M_{\text{swap}}$.



\subsection{Training Objective}
\label{sec:loss}
Different from the most current face swapping methods, our E4S takes reconstruction as the proxy task, making it easy to train.
When training is finished, one can employ the texture encoder $F_{\phi}$ to generate per-region texture codes for any input face.
Then, face swapping can be easily fulfilled as described in Sec.~\ref{sec:swappingFace}.
We apply the commonly used loss functions
in the GAN inversion methods, which are detailed in our Appendix, where the training loss of the face re-coloring network $B_{\psi}$ and face inpainting network $P_{\tau}$ are also presented.


\section{Experiment Setup}
\subsection{DataSets.}
\label{sec:dataset}
\noindent{\textbf{FaceForensics++}}~\cite{roessler2019ffplus} is a forensics dataset consisting of 1,000 original video sequences containing 1,000 human identities. The frames are mostly frontal faces without occlusions.

\noindent{\textbf{CelebAMask-HQ}}~\cite{lee2020maskgan} consists of 30K high-quality face images, which can be split into 28k and 2K for training and testing, separately. The dataset provides the facial segmentation masks, which has 19 semantic categories in total. 

\noindent{\textbf{FFHQ}}~\cite{karras2019styleGAN} contains 70K high-quality at $1024^2$
resolution. It includes vast variation in terms of age, ethnicity and image background, and also has better coverage of accessories such as eyeglasses, sunglasses, hats, etc.
Since it does not provide the facial segmentation masks, we use a pre-trained face parser~\cite{zllrunning2013faceParser} to obtain the facial segmentation masks.

\subsection{Implementation Details.}
The proposed RGI is trained with 
PyTorch~\cite{paszke2019pytorch} with 8 NVIDIA Tesla A100 GPUs. 
The batch size is 2 at each GPU during training. The learning rate is initialized as $10^{-4}$ with the Adam~\cite{kingma2014adam} optimizer ($\beta_1=0.9$, $\beta_2=0.999$). 
We train the model for 200K and 300K iterations on CelebAMask-HQ and FFHQ datasets, respectively. Besides, the initial learning rate decays by the factor of $0.1$ at 100K and 150K iterations, separately. Here, all images are randomly flipped with a ratio of $0.5$. 
As for the adversarial training, we update the parameters of the discriminator once the generator gets updated 15 times. The reason is that the pre-trained StyleGAN can well preserve the texture details of inner face components, while sometimes its performance is unstable on preserving the hair details. 
Hence, we fine-tune the first $K=13$ layers to improve the hair quality. 
As for the RGI-Optimization, we fix the learning rate as $1e^{-2}$. Empirically, approximately 50 steps of optimization would bring satisfying results.


We train the re-coloring network and inpainting network on 256$\times$256 images sampled from CelebAMask-HQ and FFHQ. Both of the networks are optimized by Adam~\cite{kingma2014adam} with $\beta_1=0.9$ and $\beta_2=0.999$.


\begin{table}[t]
\caption{Quantitative comparison of our RGI under different ablative configurations. The reconstruction performance is measured.}
\small
\vspace{-0.2cm}
\centering
\begin{tabular}{|l|cccc|}
\hline
\multicolumn{1}{|c|}{\textbf{Configurations}} & \multicolumn{1}{c}{\textbf{SSIM$\uparrow$}} & \multicolumn{1}{c}{\textbf{PSNR$\uparrow$}} & \multicolumn{1}{c}{\textbf{RMSE$\downarrow$}} & \multicolumn{1}{c|}{\textbf{FID$\downarrow$}} \\ \hline
our RGI full model                                  & 0.818                             & 19.851                            & 0.105                             & \textbf{15.032}                            \\
(A) w/o finetuning                          & 0.827                             & 19.984                            & 0.104                             & 22.239                           \\ 
(B) w/o MS encoder                        & 0.817                             & 19.732                            & 0.107                             & 15.112                            \\ \hline
\end{tabular}
\label{tbl:ablation}
\end{table}
\section{Ablation study.}
\begin{figure}[t]
\centering
\setkeys{Gin}{width=1\linewidth}
\captionsetup{belowskip=-15pt}
\begin{subfigure}{\linewidth}
\includegraphics[width=1\textwidth]{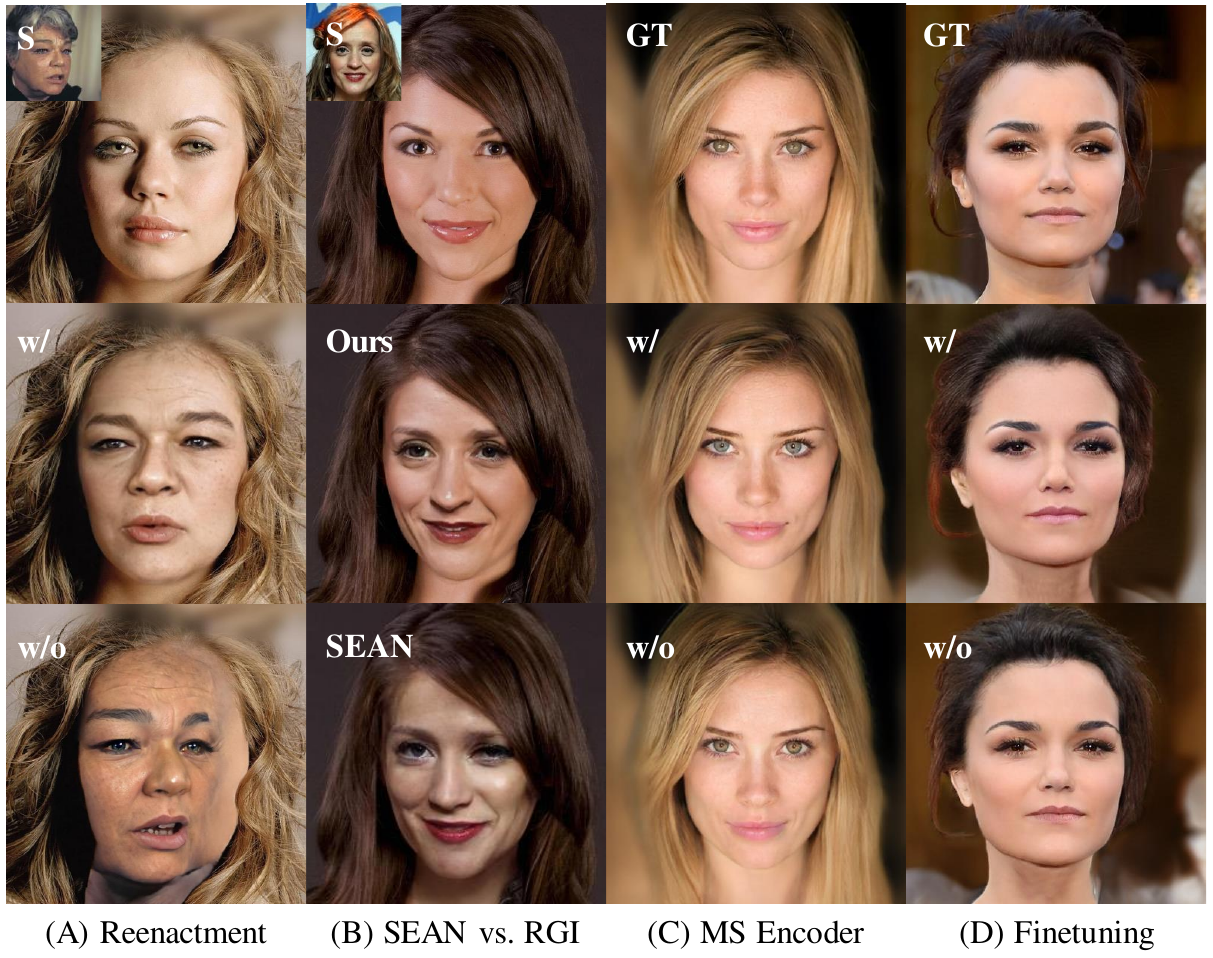}
\end{subfigure}
     \caption{Qualitative comparisons of different ablative settings. }
    \label{fig:ablation}
\end{figure}

We perform ablation studies on the different parts of the proposed E4S framework. All the ablative experiments are conducted on the CelebAMask-HQ dataset.
We provide the quantitative comparison under different RGI configurations in Table~\ref{tbl:ablation}, where the reconstruction performance is considered.  Besides, the qualitative ablation comparison in E4S is shown in Fig.~\ref{fig:ablation}.

\noindent{\textbf{The Role of Re-enactment.}}
We first study the roles of the re-enactment step of our E4S framework, which aims to drive the source to show a similar pose and expression as the target. However, aside from pose and expression, an ideal reenactment model should not affect other source attributes, such as identity.
Specifically, we employ a pre-trained face reenactment model~\cite{wang2021faceVid2Vid} before the shape and texture swapping procedure.
To verify the necessity of the re-enactment step in {E4S}, we compare a standard {E4S} pipeline and the one without re-enactment.
{As shown in the 1st column of Fig.~\ref{fig:ablation}, the swapped result is not aligned with the target face when the reenactment is disabled.}

\begin{figure*}[t]
\centering
\includegraphics[width=0.95\linewidth]{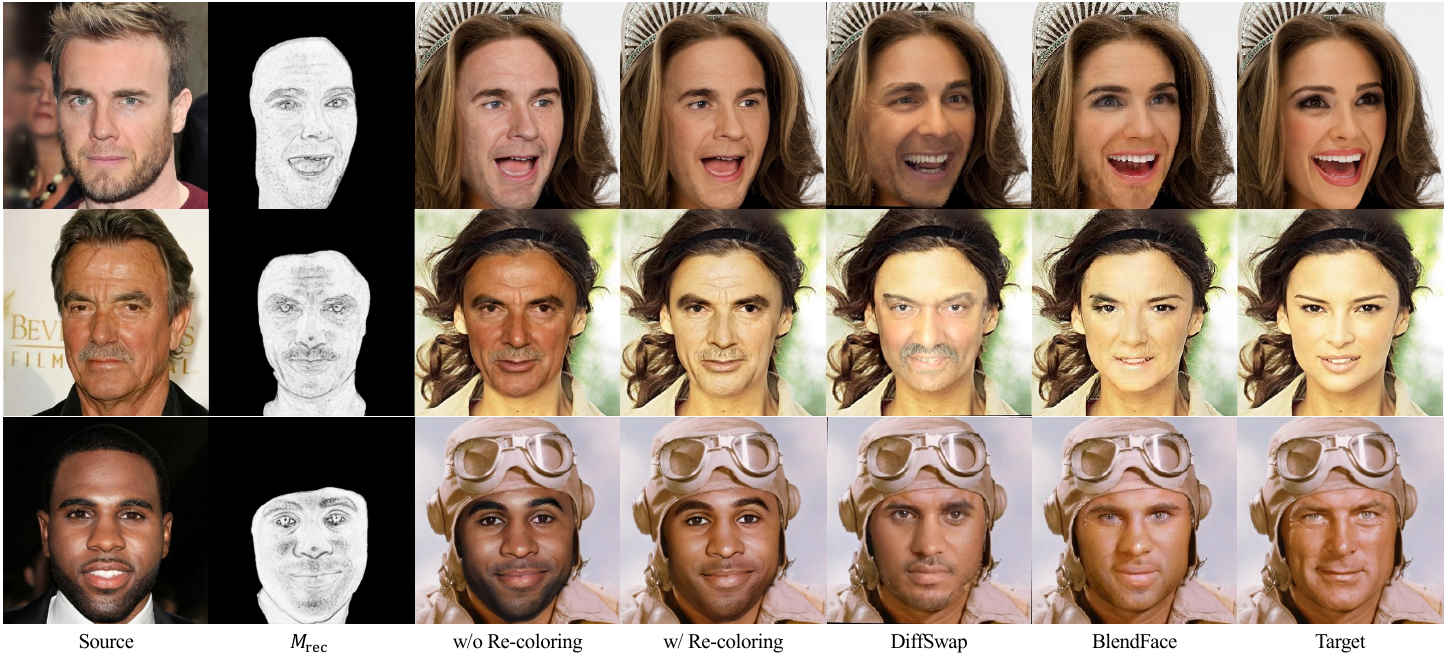}
    \captionsetup{belowskip=-6pt}
    \caption{Qualitative results of ablation study on face re-coloring. $M_{\rm{rec}}$ denotes the low-pass mask used for re-coloring. Comparing to the 3rd column ($\tilde{I}$), the results in the 4th column ($\tilde{I}_{\rm{rec}}$) have better lighting consistency with the target while preserving the skin texture of the source, demonstrating the efficacy of our re-coloring network. However, the previous methods DiffSwap~\cite{zhao2023diffswap} in the 5th column and BlendFace~\cite{shiohara2023blendface} in the 6th column fail to preserve the source identity although they keep the target lighting.
    }
    \label{fig:ablation_recolor}
\end{figure*}

\begin{figure*}[t]
\centering
\includegraphics[width=0.95\linewidth]{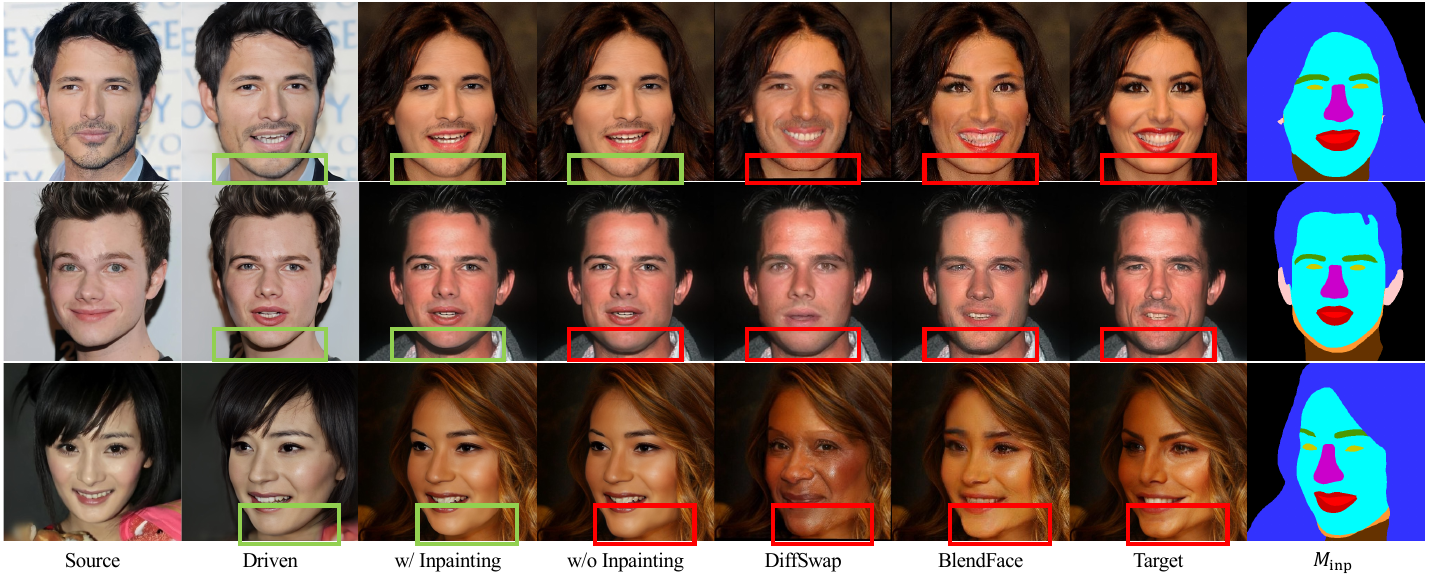}
    \captionsetup{belowskip=-7pt}
    \caption{Qualitative results of ablation study on face inpainting (please zoom in for more details). The \textcolor{green}{green} box indicates the face shape is consistent with the source image, while the \textcolor{red}{red} box means the face shape is inconsistent with the source. the inpainting segmentation map $M_{\rm{inp}}$ and the \textcolor{cssorange}{orange} pixels in $M_{\rm{inp}}$ denotes the mismatch regions $M_{\rm{r}}$.
    }
    \label{fig:ablation_inpaint}
\end{figure*}

\noindent{\textbf{SEAN \textbf{vs} RGI.}}
Our {E4S} is a general model. Specifically, if there is a method which contains an encoder extracting the per-region style codes and a generator controlling the per-region style codes along with the segmentation mask, it can be adapted to our {E4S} framework. To prove this, we replace our RGI with SEAN~\cite{zhu2020sean} to play the roles of $F_{\phi}$ and $G_{\theta}$. It can be observed from the 2nd column of Fig.~\ref{fig:ablation},  
our results preserve details better (e.g., hair texture, hair color, tone, and overall identity), whereas SEAN sometimes produces artifacts in the hair region.
Besides, SEAN only demonstrates its ability to generate faces at $256^2$ while ours are at $1024^2$. Both of these two findings show the superiority of our RGI.

\noindent{\textbf{Multi-scale vs single-scale encoder.}} We also study the role of the multi-scale encoder in configuration (B), where only the last level of feature maps produced by $F_{\phi}$ is used. Compared with our baseline, the performance of the single-scaled encoder is worse, which is consistent with the qualitative comparison shown in the 3rd column in Fig.~\ref{fig:ablation} (see the color of eyes). 
This demonstrates that the multi-scale encoder can improve the quality of generated images.

\noindent{\textbf{Pre-trained vs fine-tuned StyleGAN.}} 
The pre-trained StyleGAN can be used for face swapping. However, we notice that the hair texture details cannot be always well preserved. For a more robust performance on hair, we fine-tune the first $K=13$ layers of the StyleGAN. In Table~\ref{tbl:ablation}, we use a configuration (A) to indicate freezing the parameters of the StyleGAN generator and only training the texture encoder $F_{\phi}$ and the subsequent MLPs in our RGI. Although slightly better SSIM, PSNR, and RMSE can be achieved by (A), the FID is poor. The last column in Fig.~\ref{fig:ablation} also supports this. In contrast, fine-tuning can improve hair quality while maintaining the texture of other inner facial components. The generated images look unrealistic and lack certain textures (especially in hair). The generation quality can also be proved by the higher FID.

\noindent{\textbf{The Role of face re-coloring network $B_{\psi}$.}}
To evaluate the proposed face re-coloring network $B_{\psi}$, we provide the qualitative results in Fig.~\ref{fig:ablation_recolor}.
The comparison results demonstrate that using re-coloring network can produce more natural and target-consistent lighting.
Besides, although $B_{\psi}$ only generates $256^2$-resolution images, the final re-coloring results have the comparable image quality with the $1024^2$ sized inputs, with the help of face enhancing network~\cite{zhou2022codeformer} restoring the details and re-coloring mask $M_{\rm{rec}}$ preserving the high-frequency information from the high-resolution inputs.

\vspace{-1pt}
\noindent{\textbf{The Role of face inpainting network $P_{\tau}$.}}
We also study the role of face inpainting network $P_{\tau}$ in Fig.~\ref{fig:ablation_inpaint}. 
For clarity, we merge the mismatch region $M_{\rm{r}}$ and swapping segmentation $M_{\rm{swap}}$ into an inpainting map $M_{\rm{inp}}$ and visualize it in the last column, whose orange pixels indicate the mismatch regions.
The comparison between the 2nd and 3rd column clearly shows that the proposed network $P_{\tau}$ can successfully and adaptively keep the face shape of the source, even when the face-shape difference between source and target is extremely large. 
As shown in the 1st row, when there are no mismatch regions (no \textcolor{cssorange}{orange} pixels in $M_{\rm{inp}}$), both our methods with or without inpainting network produce the source shape consistent results. However, the state-of-the-art methods DiffSwap~\cite{zhao2023diffswap} and BlendFace~\cite{shiohara2023blendface} fail to achieve this, whose swapped faces only inherit the face shape of the target.
The results in the 2nd and 3rd rows demonstrate that when the mismatch regions (\textcolor{cssorange}{orange} pixels in $M_{\rm{inp}}$) exist, our inpainting network can restore the mismatched parts, leading to more consistent face shape with the driven sources (please zoom in for more details).

Besides, the quantitative comparison between our standard E4S and the one without face re-coloring network $B_{\psi}$ or face inpainting network $P_{\tau}$ are demonstrated in Table~\ref{tbl:swapComp}. 
We will elaborate on this in Sec.~\ref{face_swap_quantitative}.

\begin{figure*}[t]
\centering
\setkeys{Gin}{width=\linewidth}
\captionsetup{belowskip=-3pt}
\begin{subfigure}{\linewidth}
\includegraphics[width=1.0\textwidth]{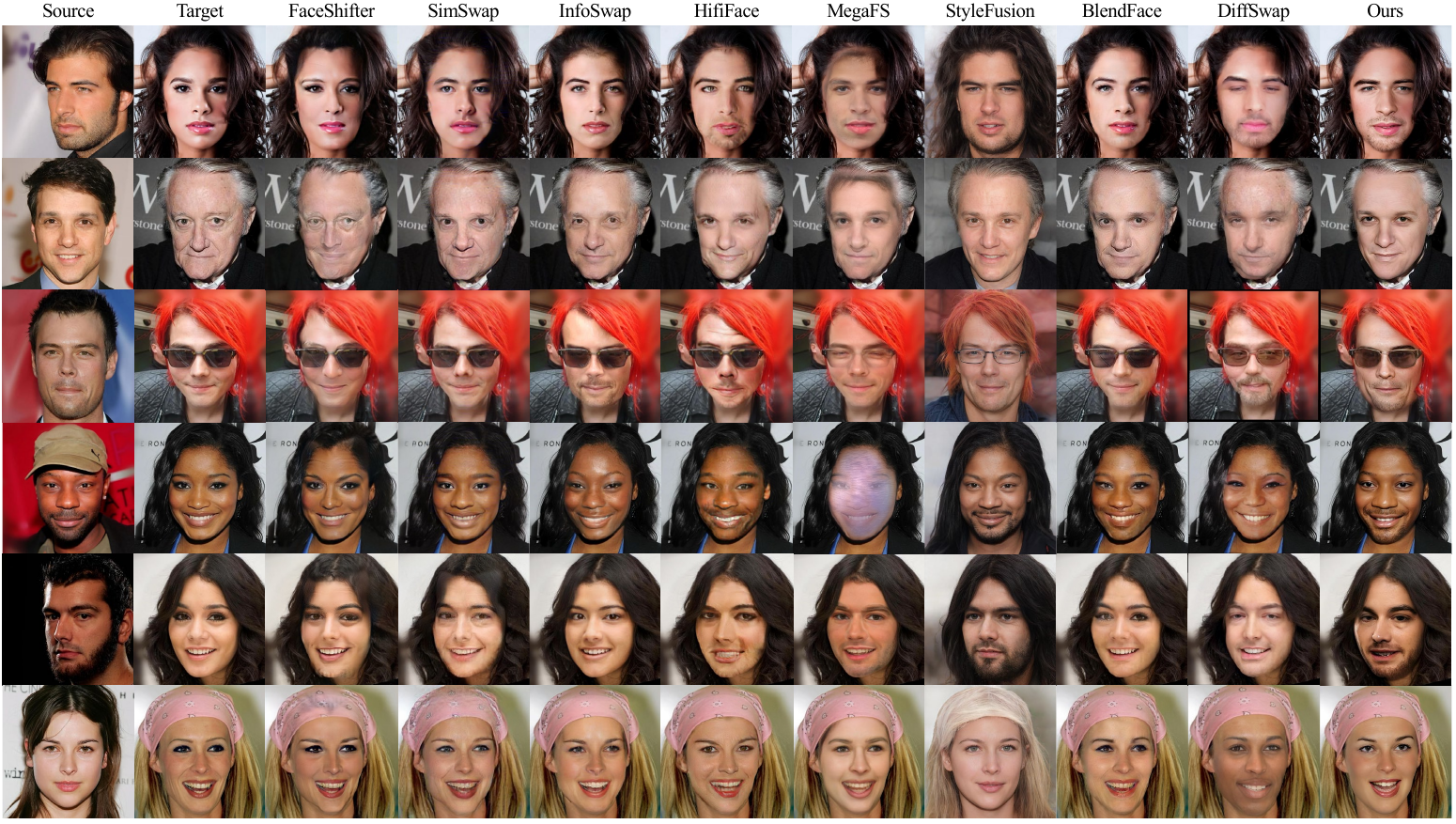}
\end{subfigure}
\caption{Qualitative comparisons of our results with state-of-the-art face swapping  methods. Our method can achieve high-fidelity results, which preserves the identity from the source better (\eg, beard, eyes, face shape) and attribute condition from the target (\eg, lighting and background). Note that our E4S maintain the skin tone from the target which is more practical in face swapping application.
Best viewed in color and zoom-in.}
\label{fig:swapComp}
\end{figure*}
\begin{figure*}[tbh]
  \centering
  \subfloat   
  {
      \label{fig:ffplus_metrics_sub1}\includegraphics[width=0.23\textwidth]{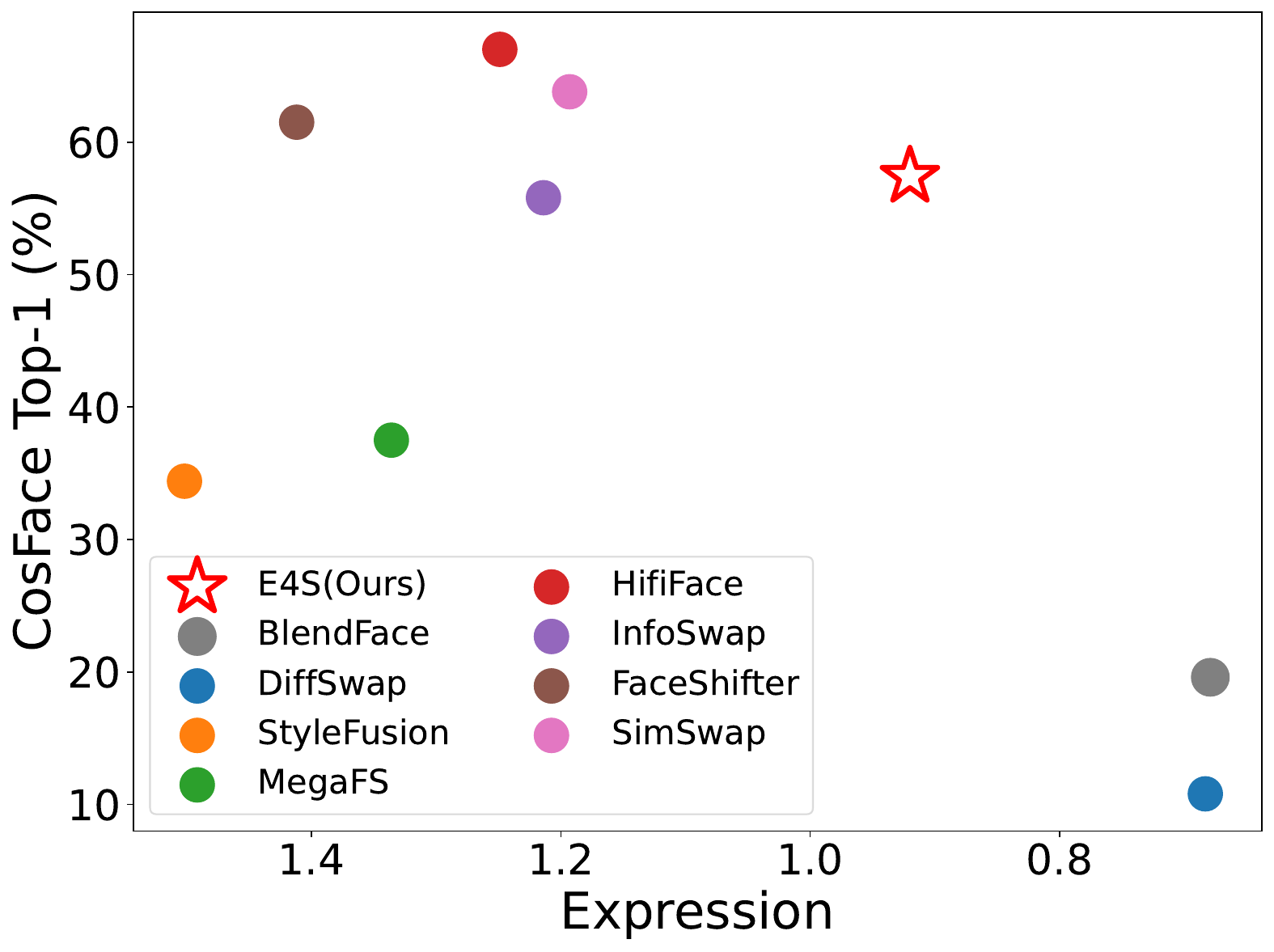}
  }
  \subfloat
  {
      \label{fig:ffplus_metrics_sub2}\includegraphics[width=0.23\textwidth]{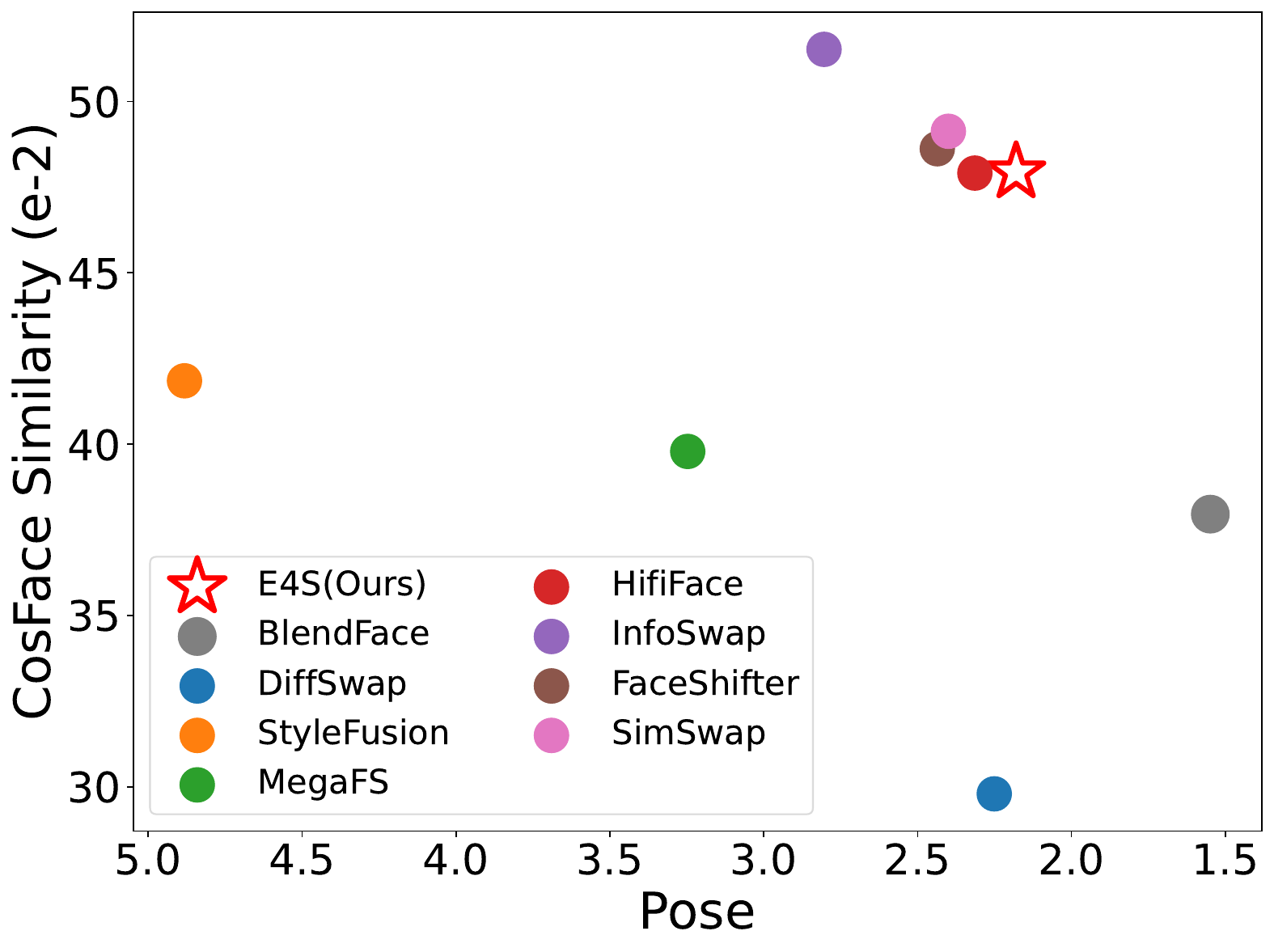}
  }
  \subfloat
  {
      \label{fig:ffplus_metrics_sub3}\includegraphics[width=0.23\textwidth]{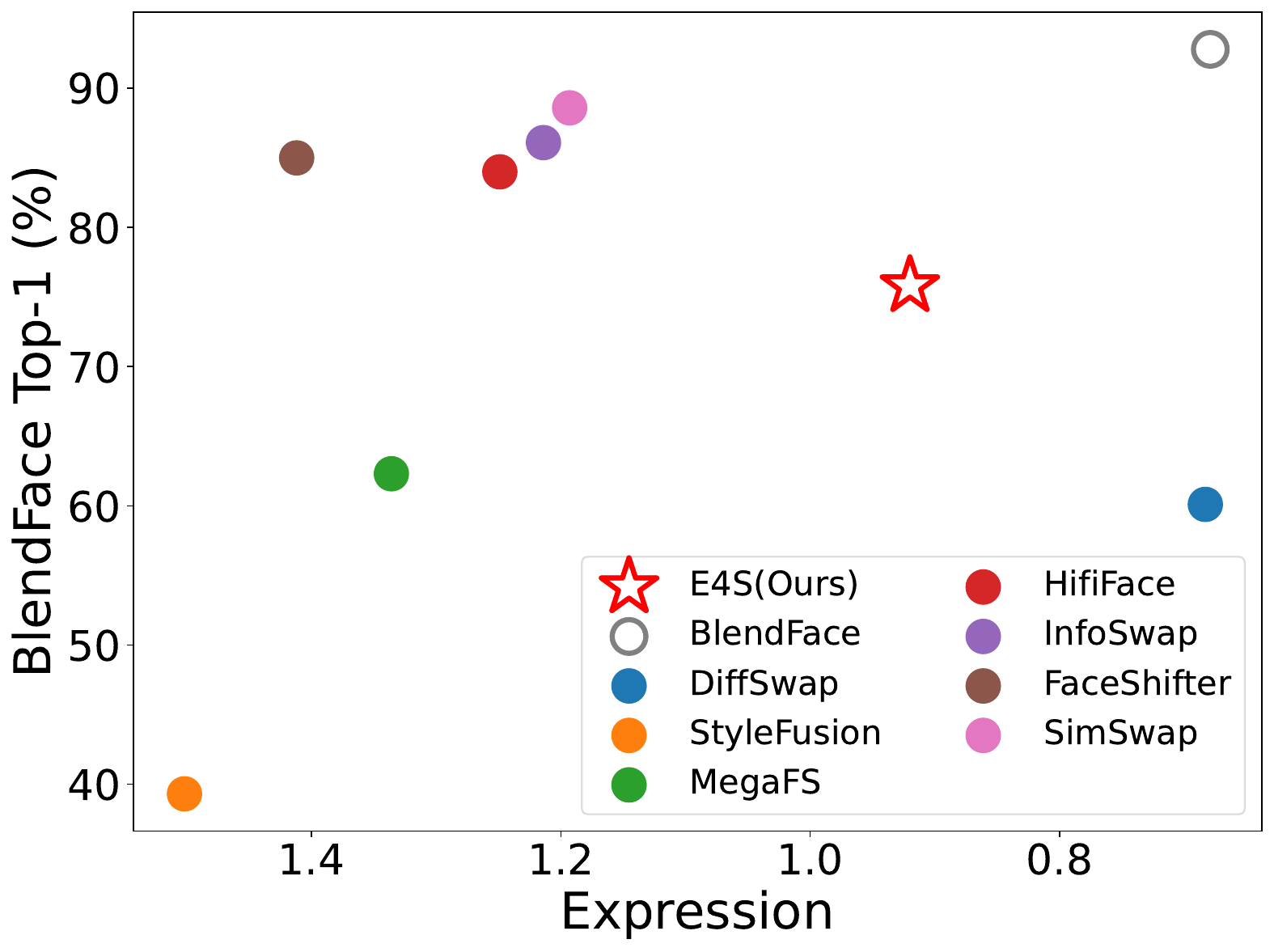}
  }
  \subfloat
  {
      \label{fig:ffplus_metrics_sub4}\includegraphics[width=0.23\textwidth]{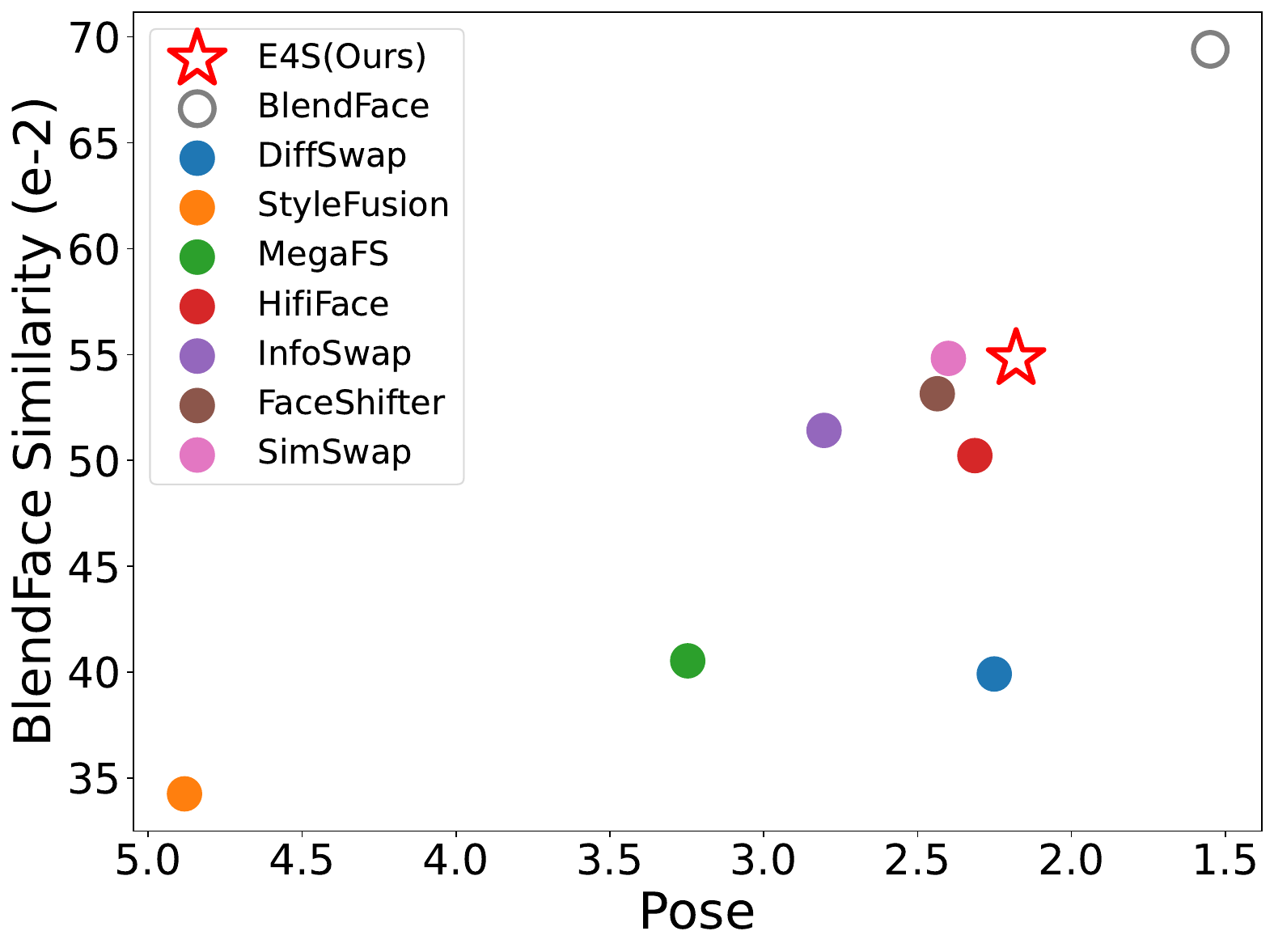}
  }
  \captionsetup{belowskip=-8pt}
  \caption{Quantitative evaluation on FaceForensics++~\cite{roessler2019ffplus}.
  The 1st and 2nd figures demonstrate that our method achieves the best trade-off between the identity similarity and target attributes preservation.
  In the 3rd and 4th figures, the gray circles indicate that BlendFace~\cite{shiohara2023blendface} uses the testing model to train the network as they stated in~\cite{shiohara2023blendface}.
  Comparing with the rest methods, our method achieve the best trade-off in the 3rd figure and the best performance in the 4th figure.
  }  
  \label{fig:ffplus_metrics}  
\end{figure*}
\begin{figure*}[tbh]
  \centering
  \subfloat   
  {
      \label{fig:celebahq_metrics_sub1}\includegraphics[width=0.23\textwidth]{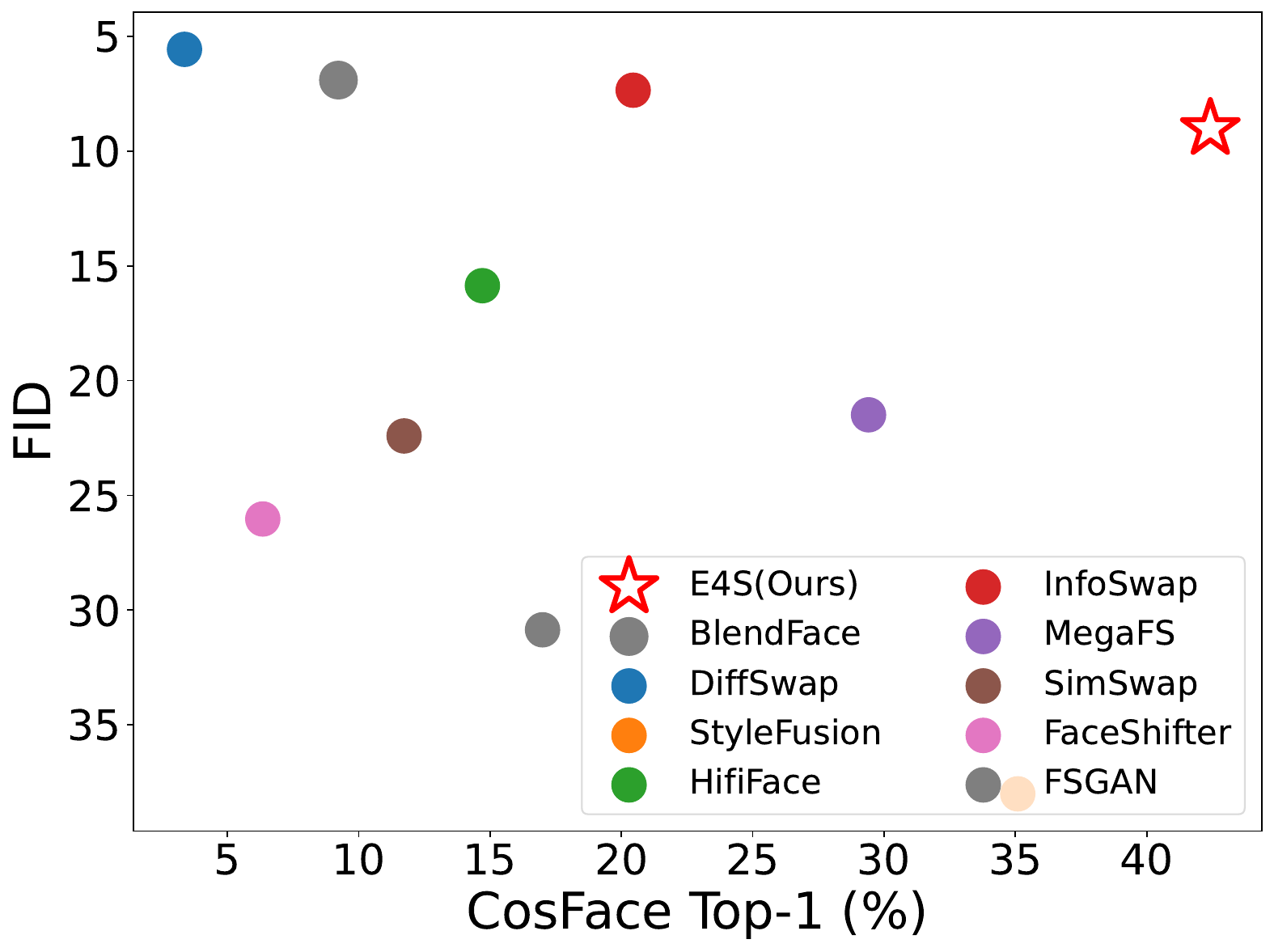}
  }
  \subfloat
  {
      \label{fig:celebahq_metrics_sub2}\includegraphics[width=0.23\textwidth]{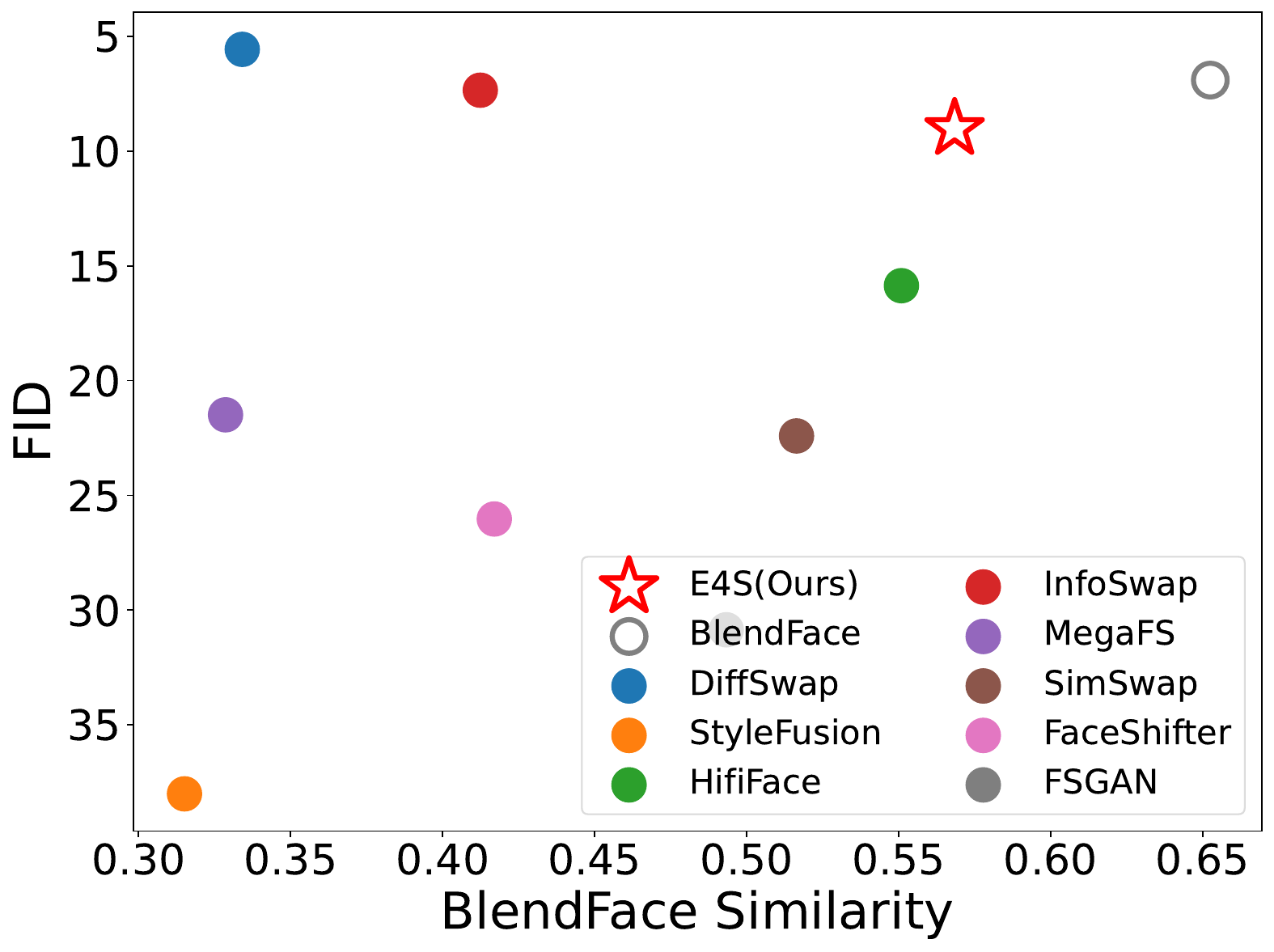}
  }
  \subfloat
  {
      \label{fig:celebahq_metrics_sub3}\includegraphics[width=0.23\textwidth]{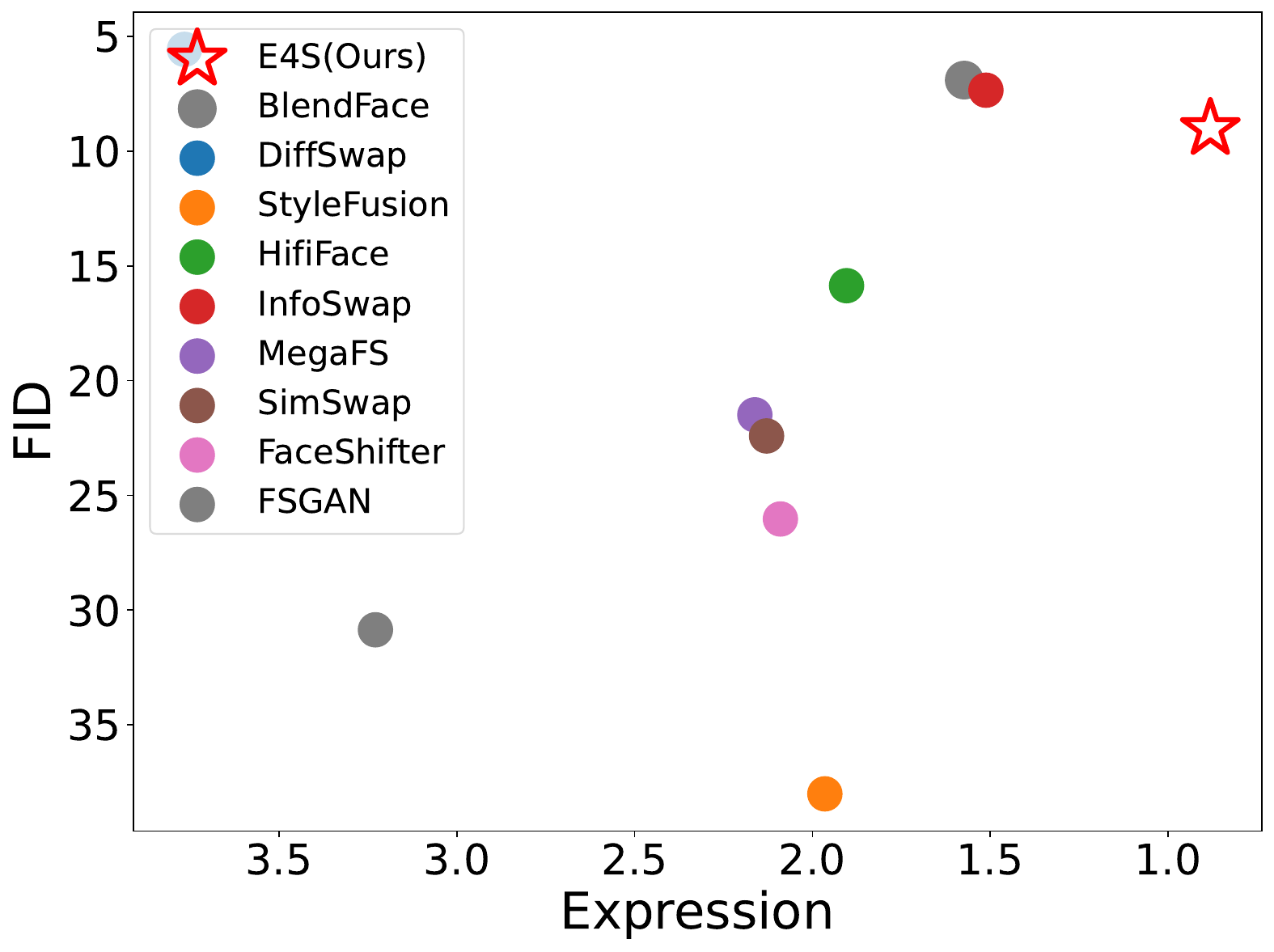}
  }
  \subfloat
  {
      \label{fig:celebahq_metrics_sub4}\includegraphics[width=0.23\textwidth]{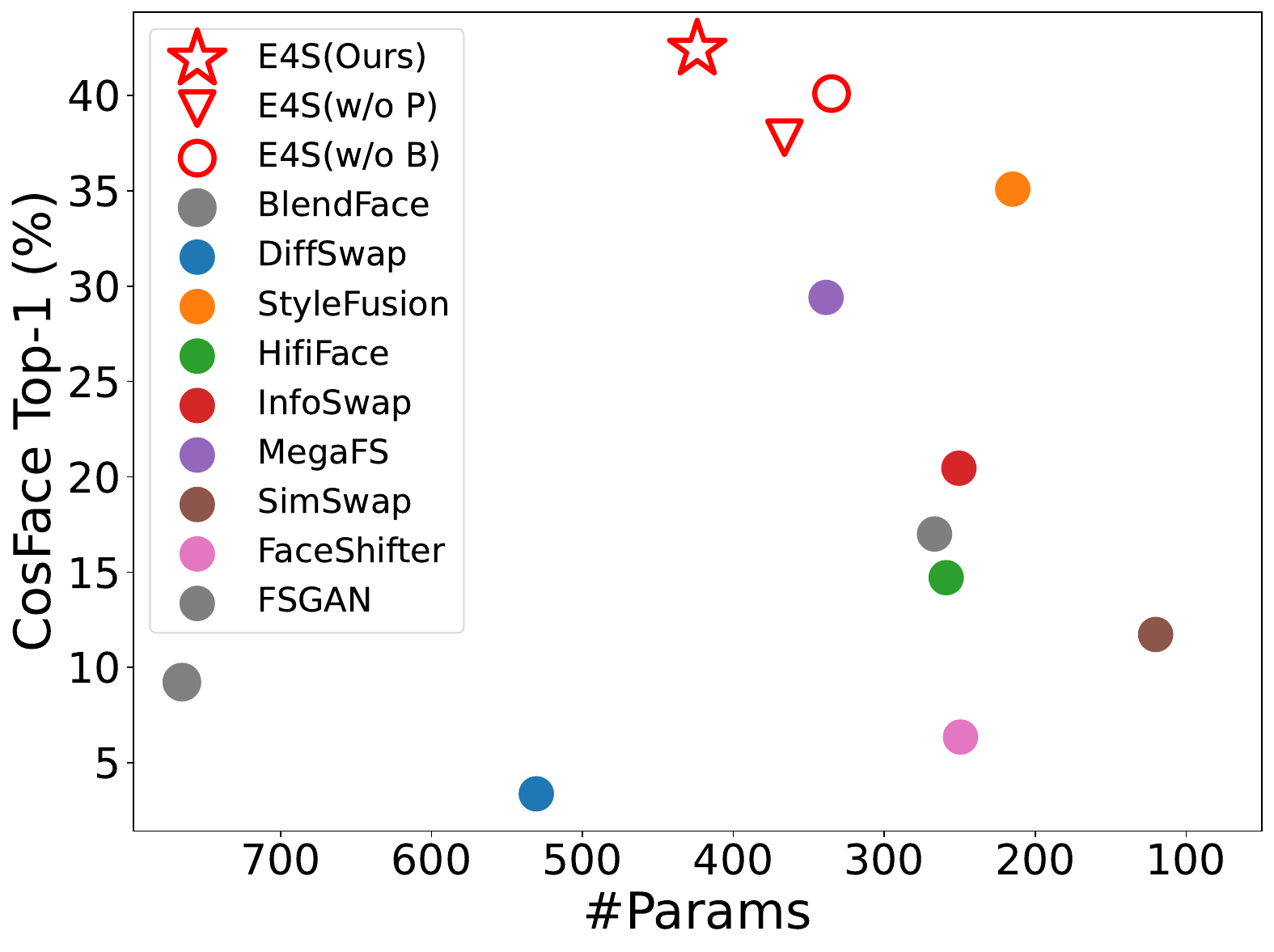}
  }
  \captionsetup{belowskip=-8pt}
  \caption{Quantitative evaluation on CelebAMask-HQ~\cite{lee2020maskgan}, where the grey circle indicates that BlendFace~\cite{shiohara2023blendface} uses the testing model to train the network, which is unfair on the BlendFace Similarity~\cite{shiohara2023blendface} metric.
  The left three figures show that our method can achieve the best trade-off between the FID and other metrics, which demonstrates our ability to natural and consistent swapped results.
  The 4-th figure shows the identity similarity and the model parameters, where our method outperforms the existing methods no matter whether it is with or without the recoloring and inpainting network.
  }  
  \label{fig:celebahq_metrics}  
\end{figure*}

\section{Face Swapping Results.}
We illustrate several face swapping examples synthesized by our E4S framework and several representative methods in Fig.~\ref{fig:swapComp}.
Our method can realize high-quality swapping at the resolution of 1024$\times$1024. 
Then, we use the 1,000 FaceForensics++ testing pairs provided by~\cite{li2019faceshifter} and randomly sample 30,000 source-target pairs from the test set of CelebAMask-HQ to obtain the quantitative results of different approaches. Besides, we provide video face swapping results in Appendix.
Moreover, we develop a user-interface system to perform image and video face swapping, which are also detailed in Appendix.


\subsection{Qualitative comparison.} 
We provide the qualitative comparisons with state-of-the-art methods in Fig.~\ref{fig:swapComp}, where our E4S achieves more realistic and high-fidelity swapped results than the others. 
The results of SimSwap~\cite{chen2020simswap} and HifiFace~\cite{wang2021hififace} suffer from some artifacts and distortions (see the 2nd and 4th row). 
Although both our E4S and FaceShifter~\cite{li2019faceshifter} can generate visually-satisfying results, ours shows better detailed textures.
InfoSwap~\cite{gao2021infoswap} fails to transfer the gender in the 1st and 4th row.

We further compare the performance in more challenging cases where the occlusion exists in the source and target faces (see the 3rd and 4th row in Fig.~\ref{fig:swapComp}). 
It can be observed that our E4S can fill out the missing skin in the source face (see the 4nd row), and preserve the target glasses (see the 3rd row).
As contrast, MegaFS~\cite{zhu2021MegaFS}, StyleFusion~\cite{kafri2021stylefusion}, BlendFace~\cite{shiohara2023blendface} and DiffSwap~\cite{zhao2023diffswap} results have many artifacts, hard to preserving the occlusions. 
The results of MegaFS~\cite{zhu2021MegaFS} look to be a mixture of the source and target, which are blurry and lack of textures.
StyleFuison~\cite{kafri2021stylefusion} shows a bit of over-smoothing (see the last row). 
BlendFace~\cite{shiohara2023blendface} and DiffSwap~\cite{zhao2023diffswap} both show very unnatural skin texture and inconsistent source identities.
As contrast, our E4S can generate more realistic and high-quality face swapping results.
The results of MegaFS~\cite{zhu2021MegaFS} look to be a mixture of the source and target, which are blurry and lack of textures.
StyleFuison~\cite{kafri2021stylefusion} shows a bit of over-smoothing (see the last row). 
As contrast, our E4S can generate more realistic and high-quality face swapping results.


\begin{figure}[t]
\centering
\setkeys{Gin}{width=\linewidth}
\begin{subfigure}{0.131\linewidth}
    \captionsetup{font={scriptsize}}
    \caption*{orig. msk}
\includegraphics{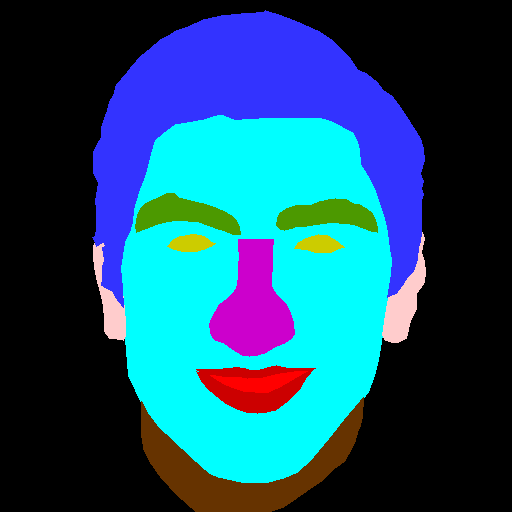}\\
\includegraphics{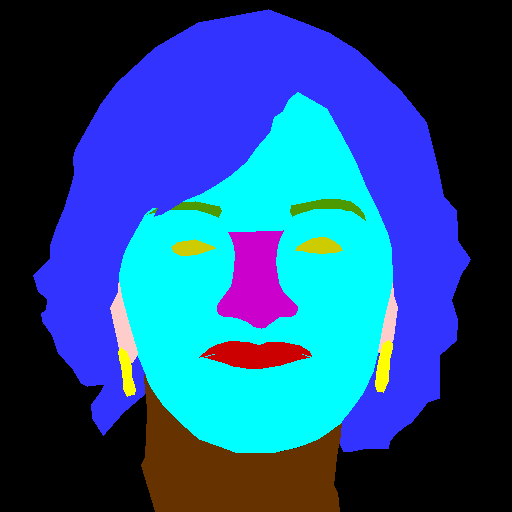}\\
\includegraphics{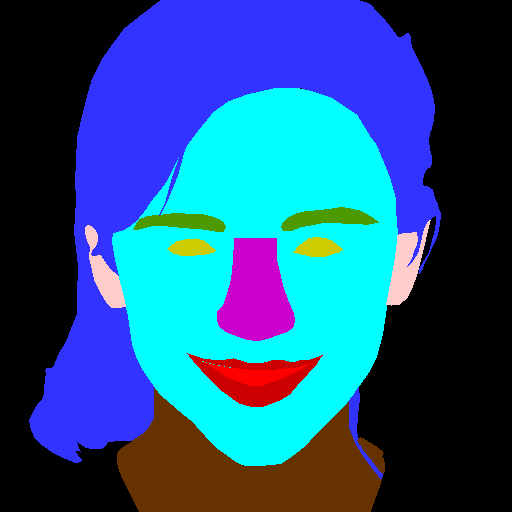}
\end{subfigure}
\begin{subfigure}{0.131\linewidth}
    \captionsetup{font={scriptsize}}
    \caption*{orig. img}
\includegraphics{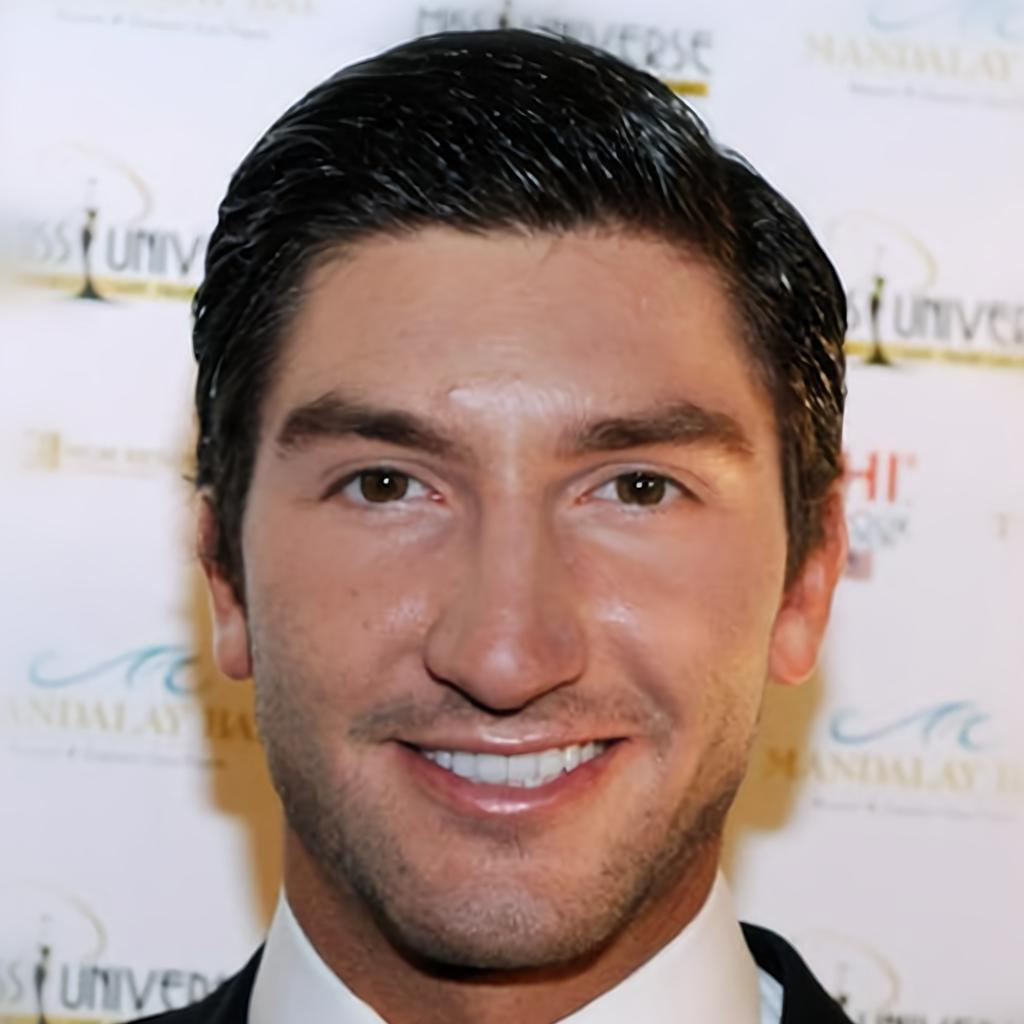}\\
\includegraphics{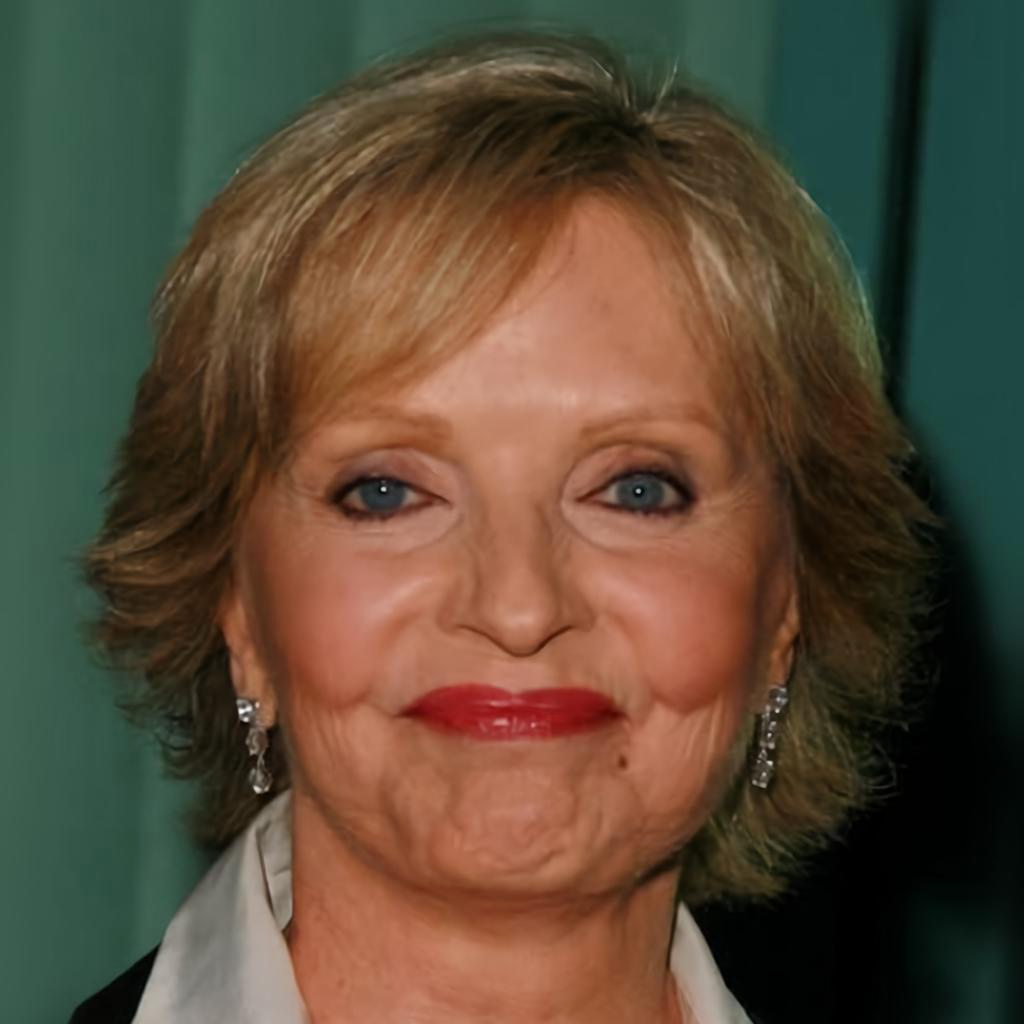}\\
\includegraphics{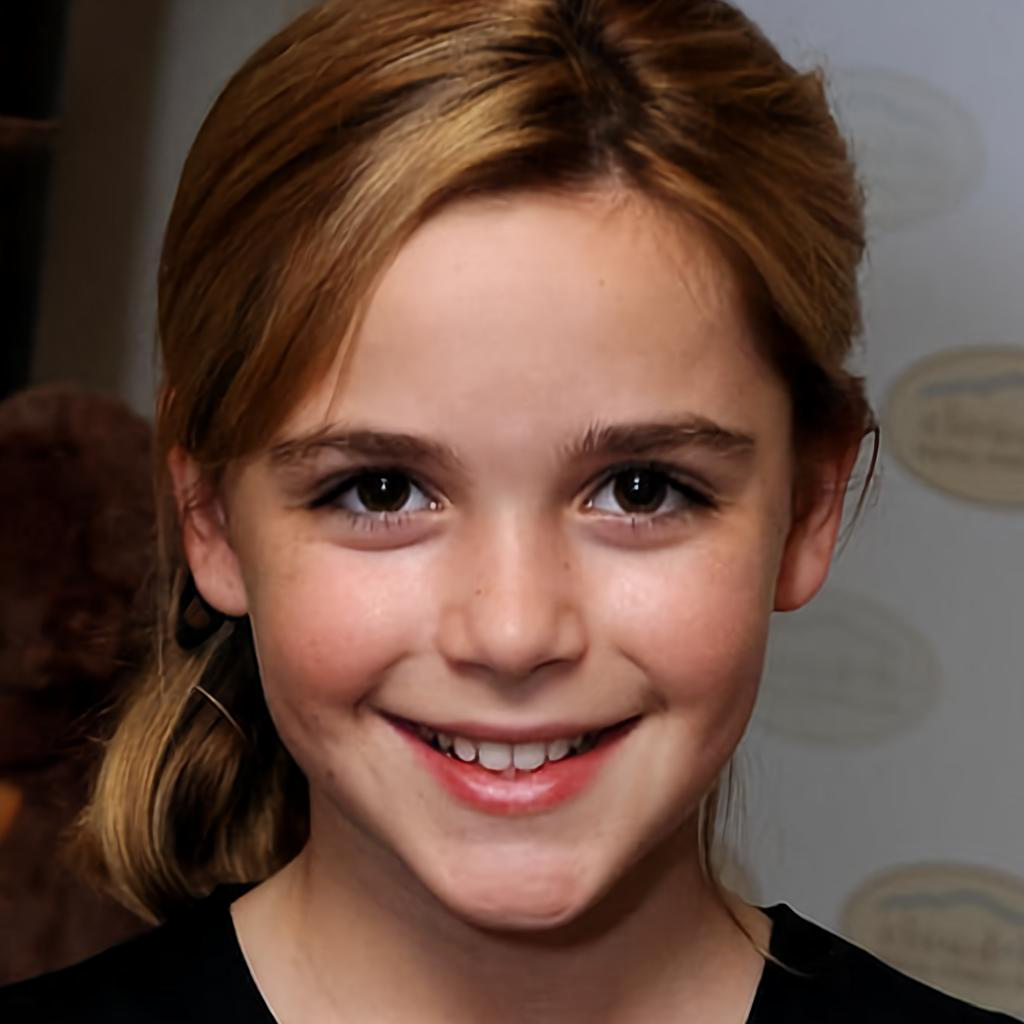}
\end{subfigure}
\begin{subfigure}{0.131\linewidth}
    \captionsetup{font={scriptsize}}
    \caption*{edited}
\includegraphics{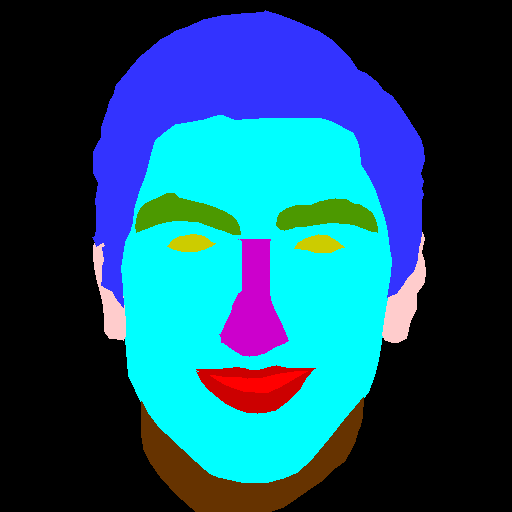}\\
\includegraphics{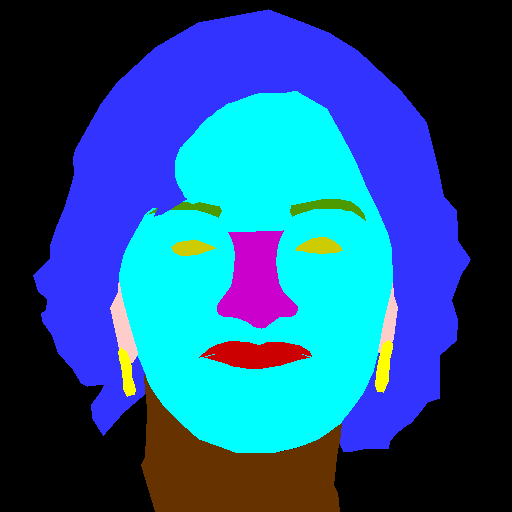}\\
\includegraphics{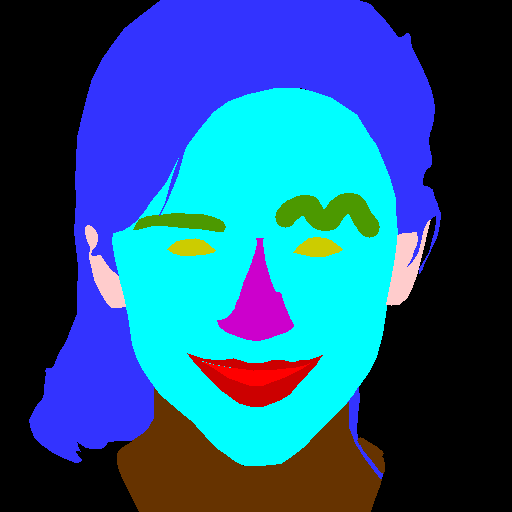}
\end{subfigure}
\begin{subfigure}{0.131\linewidth}
    \captionsetup{font={scriptsize}}
    \caption*{SPADE}
\includegraphics{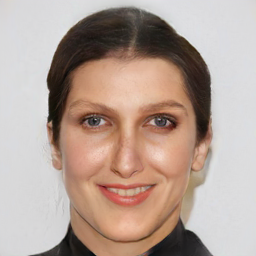}\\
\includegraphics{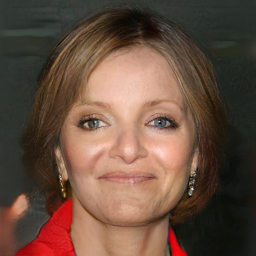}\\
\includegraphics{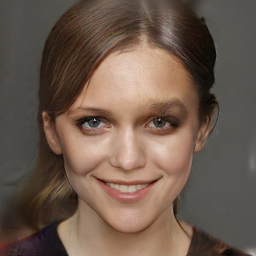}
\end{subfigure}
\begin{subfigure}{0.131\linewidth}
    \captionsetup{font={scriptsize}}
    \caption*{SEAN}
\includegraphics{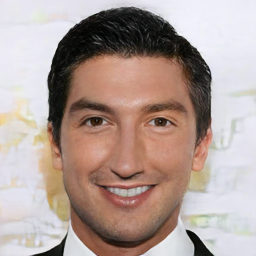}\\
\includegraphics{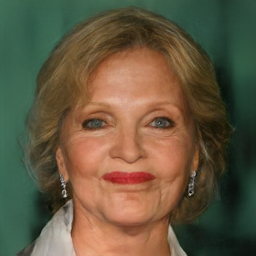}\\
\includegraphics{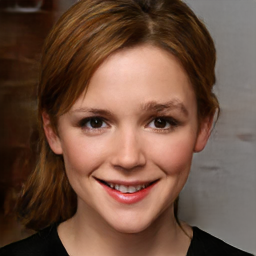}
\end{subfigure}
\begin{subfigure}{0.131\linewidth}
    \captionsetup{font={scriptsize}}
    \caption*{MaskGAN}
\includegraphics{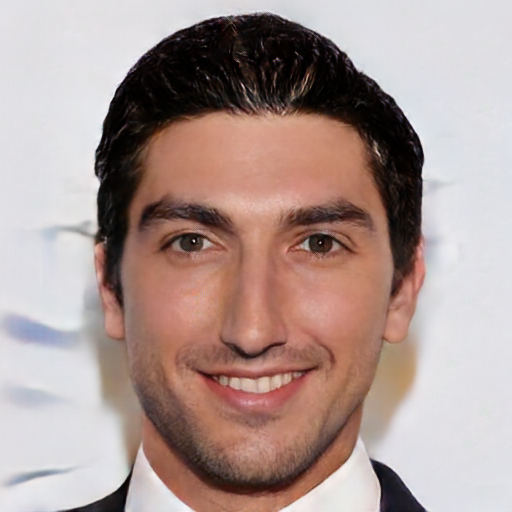}\\
\includegraphics{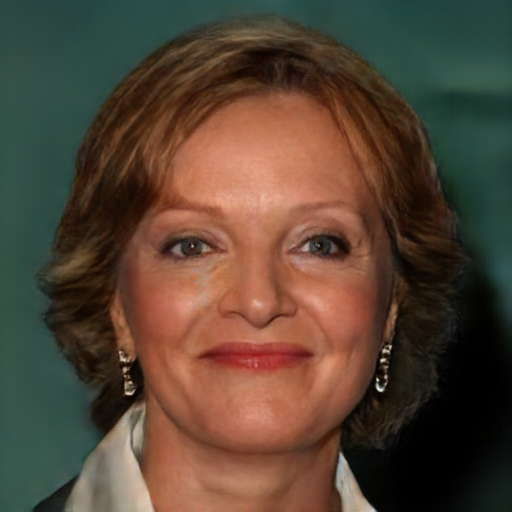}\\
\includegraphics{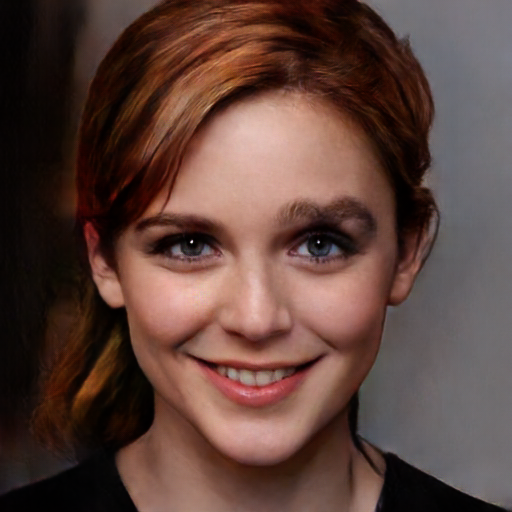}
\end{subfigure}
\begin{subfigure}{0.131\linewidth}
    \captionsetup{font={scriptsize}}
    \caption*{Ours}
\includegraphics{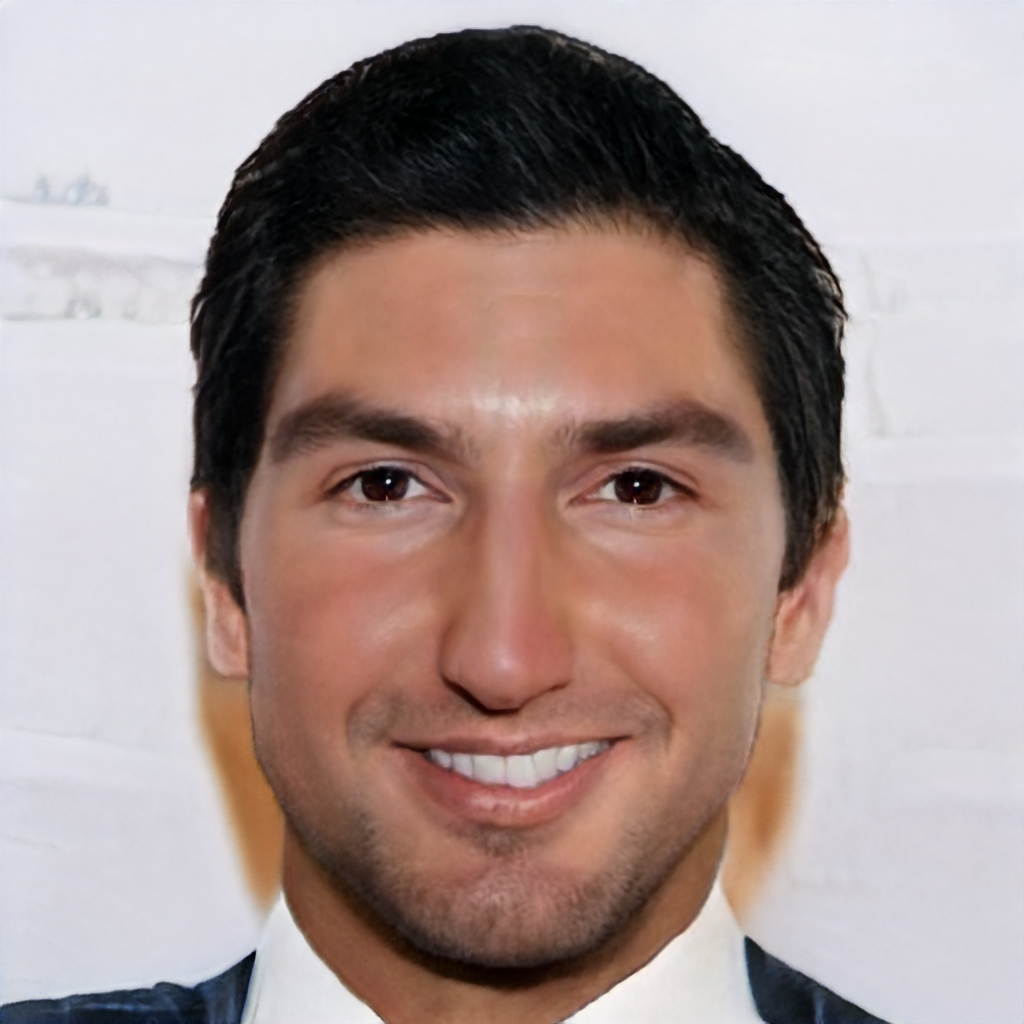}\\
\includegraphics{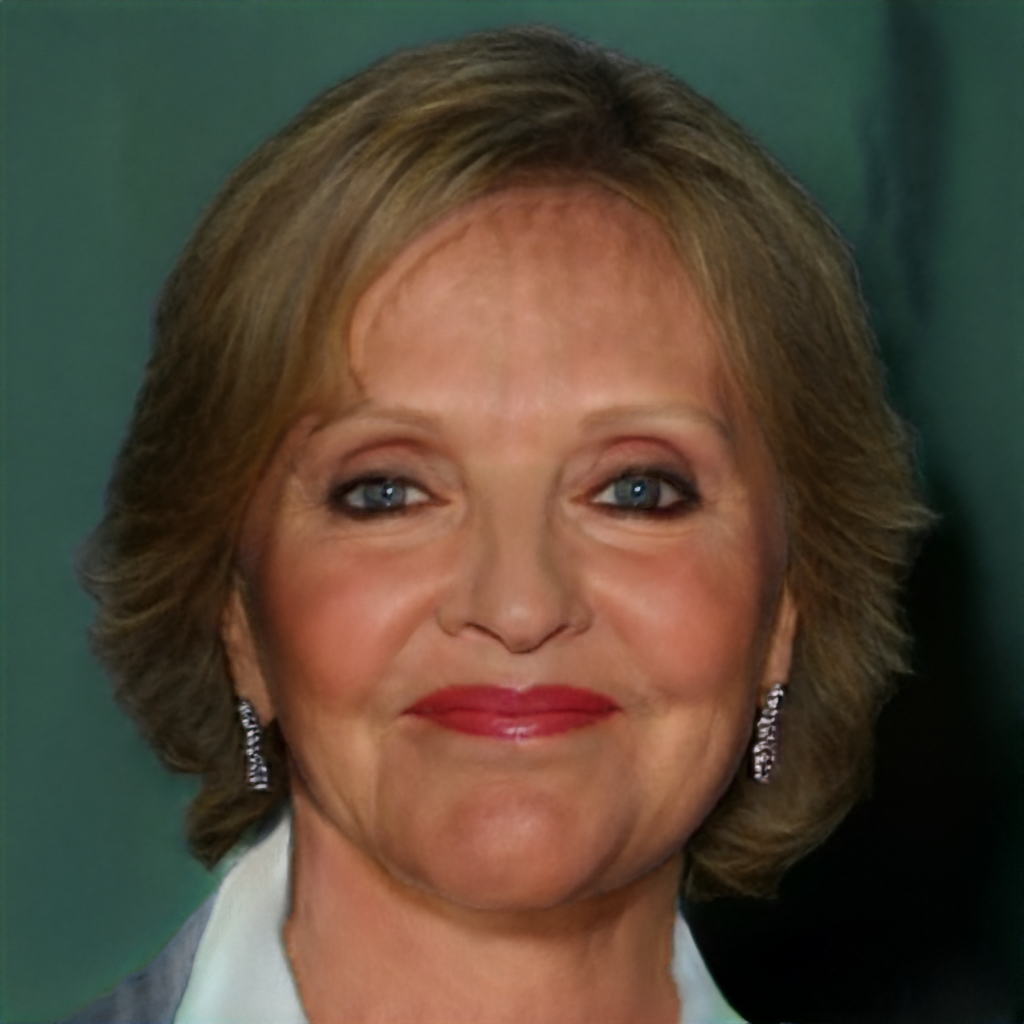}\\
\includegraphics{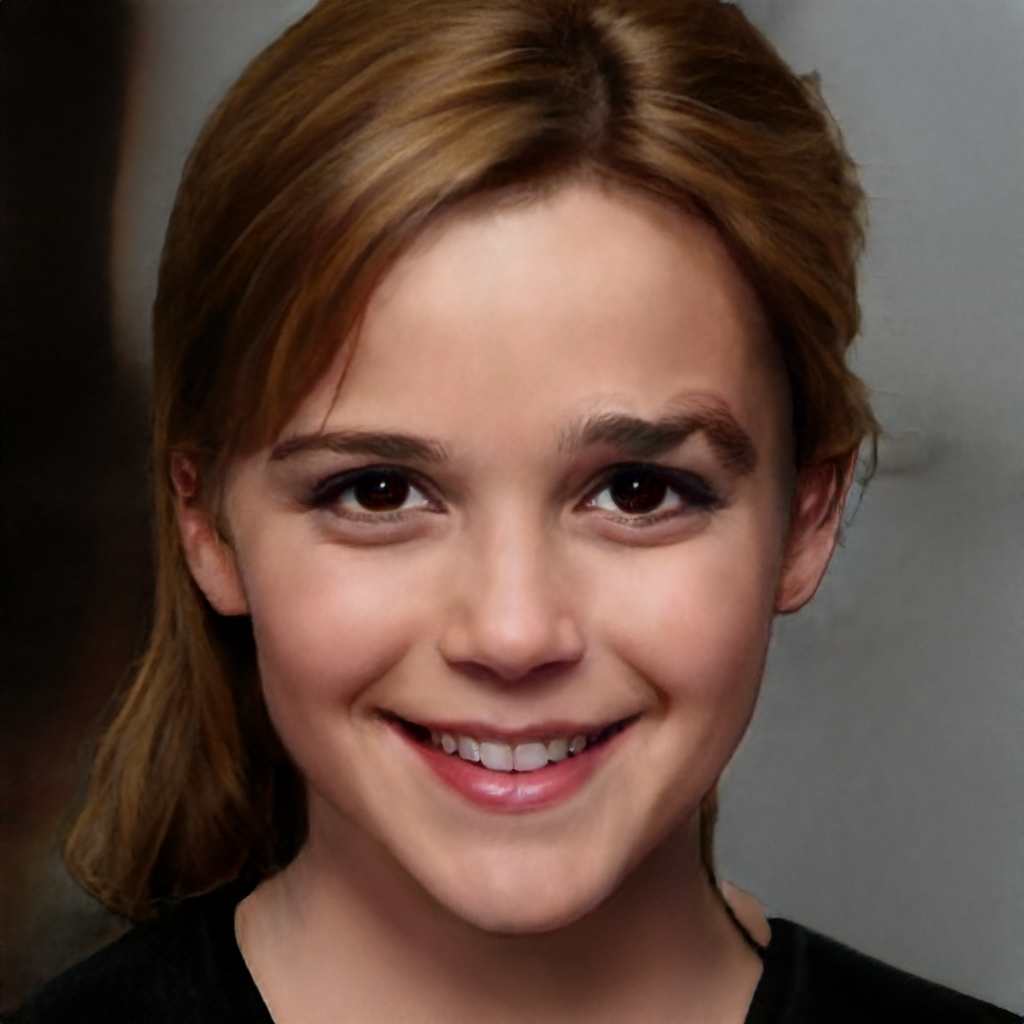}
\end{subfigure}
    \captionsetup{belowskip=-15pt}
    \caption{Qualitative comparisons with state-of-the-art face editing methods. Some modifications are made on the face contour, hair, nose, and eyebrows. Our method can produce more high-fidelity editing results while maintaining the details of other components and the overall identity information well.}
    \label{fig:edit}
    \end{figure}

\begin{table*}[t]
\centering
\caption{Quantitative comparison with the existing methods on CelebAMask-HQ testing dataset~\cite{lee2020maskgan}. We denote the best results with \textbf{bold} marks, the 2nd and 3rd highest scores are noted with $\underline{\rm{underline}}$ and $\dotuline{\rm{dot}}$ respectively. Besides, the gray numbers mean that the method uses the testing model to train the network, resulting in high BlendFace-related scores~\cite{shiohara2023blendface}.
}
\label{tbl:swapComp}
\begin{tabular}{l|ccc|cc|c|c|c|cc} 
\toprule
\multirow{2}{*}{\textbf{Method}}                         & \multicolumn{3}{c|}{\textbf{CosFace$\uparrow$}~}  & \multicolumn{2}{c|}{\textbf{BlendFace$\uparrow$}} & \multirow{2}{*}{\textbf{Pose$\downarrow$}} & \multirow{2}{*}{\textbf{Expr.$\downarrow$}} & \multirow{2}{*}{\textbf{FID\textbf{$\downarrow$}}} & \multirow{2}{*}{\textbf{Res}.} & \multirow{2}{*}{\textbf{\#Params}}  \\
                                                         & Top-1(\%)      & Top-5(\%)      & Sim.             & Top-1(\%)     & Sim.                              &                                            &                                             &                                                    &                                &                                     \\ 
\midrule
FSGAN~\cite{nirkin2019fsgan}           & 17.00          & 32.18          & 0.2691          & 63.40         & 0.4933                            & \dotuline{2.332}                              & 3.229                                       & 30.86                                              &     $256^2$        & 266.72                              \\
FaceShifter~\cite{li2019faceshifter}    & 6.35           & 31.77          & 0.3161          & 48.95         & 0.4170                            & \textbf{1.731}                             & 2.090                                       & 26.03                                              &        $256^2$       & 249.50                              \\
SimSwap~\cite{chen2020simswap}          & 11.73          & 26.09          & 0.3665          & $\underline{75.68}$         & \dotuline{0.5164}                    & 2.892                                      & \textbf{2.129}                              & 22.41                                              &       $256^2$      & 120.21                              \\
InfoSwap~\cite{gao2021infoswap}                                                 & 20.45          & 50.20          & 0.3284          & 47.56         & 0.4124                            & $\underline{2.218}$                                      & 1.512                                       & \dotuline{7.34}                                       &         $512^2$          & 250.56                              \\
MegaFS~\cite{zhu2021MegaFS}             & 29.41          & 45.48          & 0.2588          & 31.59         & 0.3286                            & 3.044                                      & 2.162                                       & 21.49                                              &       $1024^2$        & 338.60                              \\
HifiFace~\cite{wang2021hififace}        & 14.71          & 36.94          & 0.3656          & \textbf{81.70}         & $\underline{0.5509}$                            & 2.768                                      & 1.904                                       & 15.86                                              &        $256^2$   & 258.99                              \\
StyleFusion~\cite{kafri2021stylefusion} & 35.09          & 48.76          & 0.3206          & 16.89         & 0.3151                            & 6.053                                      & 1.965                                       & 38.01                                              &      $1024^2$     & 214.89                              \\
DiffSwap~\cite{zhao2023diffswap}                       & 3.37           & 24.61          & 0.1947          & {30.29}         & 0.3341                            & 3.183                                      & 3.766                                       & \textbf{5.56}                                      &           $256^2$ & 530.57                              \\
BlendFace~\cite{shiohara2023blendface}                        & 9.23           & 35.84          & 0.2784          & \textcolor{mygray}{91.03}         & \textcolor{mygray}{0.6525}                            & 2.976                                      & 1.573                                       & $\underline{6.90}$                                               &        $256^2$   & 765.40                              \\ 
\midrule
E4S (\textit{w/o} $B_\psi$)                                & $\underline{40.10}$          & $\underline{57.24}$         & $\underline{0.3716}$          & 73.10         & 0.5043                            & 3.287                                      & \dotuline{0.959}                               & 10.17                                              &            $1024^2$  & 334.95                              \\
E4S (\textit{w/o} $P_\tau$)                                & \dotuline{37.78}  & \dotuline{56.98}  & \dotuline{0.3684}  & 74.55         & 0.4982                            & 3.275                                      & $\underline{0.910}$                                       & 9.04     &      $1024^2$      & 366.13                              \\
E4S (Ours)                                                     & \textbf{42.42} & \textbf{58.22} & \textbf{0.3829} & \dotuline{75.02} & \textbf{0.5684}                   & 3.281                                      & \textbf{0.881}                              & 9.02                                               & $1024^2$                       & 423.94                              \\
\bottomrule
\end{tabular}
\end{table*}

\subsection{Quantitative comparison.} 
\label{face_swap_quantitative}
The quantitative comparison is not only illustrated in Fig.~\ref{fig:ffplus_metrics} and Fig.~\ref{fig:celebahq_metrics} to clearly show the tradeoff between different metrics, but also detailed in Table~\ref{tbl:swapComp}.
Note that in Fig.~\ref{fig:ffplus_metrics}, we empirically found that the 10,000 testing pairs of FaceForensics++~\cite{roessler2019ffplus} having 10 images per identities can lead to pretty high scores (usually over 99\%), making it hard to distinguish the performance of the existing methods.
Therefore, we chose to use only one image per identity, which is a harder testing protocol, leading to the lower but more distinguished scores.
The evaluation protocol mainly considers two aspects: \textit{identity preservation} from the source and the \textit{attribute preservation} from the target. 
Specifically, inspired by FaceShifter~\cite{li2019faceshifter}, our experimental settings are as follows. For source identity preservation, we first extract the ID feature vectors of all the source faces and the swapped results via CosFace~\cite{wang2018cosface}. 
Then, we conduct face retrieval by searching for the most similar face from all the source images for each swapped face, where the similarity is computed as the cosine distance.
We use Top-1 and Top-5 CosFace~\cite{wang2018cosface} retrieval accuracy, cosine similarity and Top-1 BlendFace~\cite{shiohara2023blendface} retrieval accuracy and consine similarity as the source identity consistency metrics.
For the target attribute preservation, we estimate the pose and expression by using HopeNet~\cite{Ruiz2018HopeNet} and a 3D face reconstruction model~\cite{deng2019accurate}, respectively. 
The $\ell_2$ distance of the pose and expression between each swapped face and its ground-truth target face is used as metrics.
We also calculate the FID~\cite{heusel2017FID} score to compare the image quality of the methods.

Table~\ref{tbl:swapComp} shows our E4S achieves the best retrieval accuracy, which demonstrates the superiority of our method on identity preservation. As for the target attribute preservation, our E4S achieves comparable results in pose and expression. In general, the target-oriented methods~\cite{li2019faceshifter, chen2020simswap, gao2021infoswap, shiohara2023blendface} usually keep more accurate pose and expression, since they start from the target face while our E4S is a source-oriented method, which generates the target attributes starting from the source.  Nonetheless, one side effect of the target-oriented methods is that the injected identity information from the source is always limited (see FaceShifter and BlendFace in Fig.~\ref{fig:swapComp}).
To sum up, there is a trade-off between identity and attribute preservation in source- and target-oriented methods. Note that our E4S has the potential to achieve better results by leveraging a more advanced reenactment method.

Besides, we also evaluate how the proposed face re-coloring network $B_{\psi}$ and face inpainting network $P_{\tau}$ make a difference to the overall E4S framework in Table~\ref{tbl:swapComp}. Although the face re-coloring network $B_{\psi}$ basically does not affect the performance on source ID retrieval and target pose and expression, it totally changes the visual effects of human eyes since the face skin tone is maintained from the target rather the source (see Fig.~\ref{fig:ablation_recolor}).
Besides, Fig.~\ref{fig:ablation_inpaint} has demonstrated our face inpainting network $P_{\tau}$ can successfully maintain the face shape no matter the mismatch regions exist (2nd and 3rd rows) or not (1st row),
which cannot be solved well by mask-swapping only. 
The quantitative comparison in Table~\ref{tbl:swapComp} demonstrates our face inpainting network $P_{\tau}$ can improve the performance of ID retrieval since accurately maintaining source face shape is important for source ID retrieval. Such a performance improvement is consistent with the results in Fig.~\ref{fig:ablation_inpaint}.

\section{Face Editing Results Via RGI.}
Other than face swapping, our RGI can also be used for fine-grained face editing.
Here, we make a comparison with our method and the current state-of-the-art fine-grained face editing works: SPADE~\cite{park2019semantic}, Mask-guided GAN~\cite{gu2019mask}, SEAN~\cite{zhu2020sean}, and MaskGAN~\cite{lee2020maskgan}. 
To present a fair comparison, we train our RGI on the training set of CelebAMask-HQ and evaluate it on the test set.

\noindent{\textbf{{Qualitative comparison.}}
We show the visual comparison with the competing methods in Fig.~\ref{fig:edit}. We make some modifications to the original facial mask, such as hair, eyebrows, and face contour. 
It shows that our approach produces more high-fidelity editing results, where the details of other components and the overall identity information are well kept.

 \begin{table}[t]
 \begin{center}
\caption{Quantitative comparison for image reconstruction on CelebAMask-HQ~\cite{lee2020maskgan} test set. The rows in \colorbox{mygray}{gray} indicate the reconstructions are obtained via style code optimization.}
\label{tab:recon}
\begin{tabular}{l|c|c|c|c}
\toprule 
\multicolumn{1}{c|}{\textbf{Method}} & \textbf{SSIM$\uparrow$} & \textbf{PSNR$\uparrow$} & \textbf{RMSE$\downarrow$} & \textbf{FID$\downarrow$} \\ \hline
SPADE~\cite{park2019semantic}                           & 0.64       & 15.67    & 0.17        & 20.45        \\
SEAN~\cite{zhu2020sean}          & 0.71                    & 18.57  & 0.12       & 17.74        \\
MaskGAN~\cite{lee2020maskgan}    & 0.75                    & 19.42  & 0.11       & 19.03        \\ 
Our RGI & \textbf{0.82}         & \textbf{19.85}  & \textbf{0.10}       & \textbf{15.03}        \\
\hline
\rowcolor{mygray} sofGAN~\cite{chen2021sofgan}  & 0.76            & 14.86   & 0.19  & 26.73          \\ 
\rowcolor{mygray} RGI-Optimization  & \textbf{0.86}  & \textbf{23.02} & \textbf{0.07} &\textbf{14.73}        \\ \bottomrule
\end{tabular}
\end{center}
\end{table}
\noindent{\textbf{{Quantitative comparison.}}
We make comparison between the competing methods and our RGI on the image reconstruction quality, where the Structural Similarity Index (SSIM)~\cite{wang2004SSIM}, Root-Mean-Squared-Error (RMSE), Peak Signal-to-Noise Ratio (PSNR), and Fréchet Inception Distance (FID)~\cite{heusel2017FID} are used as the metrics. The results are showed in Table~\ref{tab:recon}.
We also compare E4S with  SofGAN~\cite{chen2021sofgan}, which is a StyleGAN-based method and takes style code optimization for the reconstruction. For a fair comparison, we use RGI-Optimization. 
Table~\ref{tab:recon} shows that our method always achieves the best performance on all metrics, indicating the superiority of our method. 
SEAN~\cite{zhu2020sean} sometimes produces artifacts on hair regions while our RGI can achieve more high-fidelity reconstructions with better identity, texture, and illumination. 
Besides, our RGI-Optimization can preserve the facial details better (\eg, the curly degree of hair, the thickness of the beard, dimples, and background).

\section{Conclusion}
In this paper, we present a novel framework E4S for face swapping, which considers conducting face swapping from the perspective of fine-grained face editing, \ie, \textit{``editing for swapping''}. 
Our E4S proposes to explicitly disentangle the shape and texture of each facial component, thus the face swapping task can be reformulated as a simplified problem of texture and shape swapping. To achieve such disentanglement as well as high resolution and high fidelity, we propose a novel Regional GAN inversion (RGI) method. 
Specifically, a multi-scale mask-guided encoder is designed to project the input face into the per-region style codes, which are resident in the style space of StyleGAN. Besides, we design a mask-guided injection module that uses the style codes to manipulate the feature maps in the generator according to the given masks. Besides of shape of texture, we propose to transfer the target lighting to the swapped face by training a face re-coloring network, which is trained in a self-supervised manner to recover the corresponding grayscale images. Further, we design a face inpainting network to maintain the source face shape on the pixel level. We conduct extensive experiments on face swapping, face editing and some extended applications. The results and comparisons with current state-of-the-art methods demonstrate the superiority of the E4S framework, RGI method, face re-coloring and face inpainting network.

\normalem
\bibliographystyle{IEEEtran}
\bibliography{ref}

\begin{thebibliography}{10}
\providecommand{\url}[1]{#1}
\csname url@samestyle\endcsname
\providecommand{\newblock}{\relax}
\providecommand{\bibinfo}[2]{#2}
\providecommand{\BIBentrySTDinterwordspacing}{\spaceskip=0pt\relax}
\providecommand{\BIBentryALTinterwordstretchfactor}{4}
\providecommand{\BIBentryALTinterwordspacing}{\spaceskip=\fontdimen2\font plus
\BIBentryALTinterwordstretchfactor\fontdimen3\font minus \fontdimen4\font\relax}
\providecommand{\BIBforeignlanguage}[2]{{%
\expandafter\ifx\csname l@#1\endcsname\relax
\typeout{** WARNING: IEEEtran.bst: No hyphenation pattern has been}%
\typeout{** loaded for the language `#1'. Using the pattern for}%
\typeout{** the default language instead.}%
\else
\language=\csname l@#1\endcsname
\fi
#2}}
\providecommand{\BIBdecl}{\relax}
\BIBdecl

\bibitem{zhu2021MegaFS}
Y.~Zhu, Q.~Li, J.~Wang, C.-Z. Xu, and Z.~Sun, ``One shot face swapping on megapixels,'' in \emph{Proceedings of the IEEE/CVF Conference on Computer Vision and Pattern Recognition}, 2021, pp. 4834--4844.

\bibitem{kafri2021stylefusion}
O.~Kafri, O.~Patashnik, Y.~Alaluf, and D.~Cohen-Or, ``Stylefusion: A generative model for disentangling spatial segments,'' \emph{arXiv preprint arXiv:2107.07437}, 2021.

\bibitem{shiohara2023blendface}
K.~Shiohara, X.~Yang, and T.~Taketomi, ``Blendface: Re-designing identity encoders for face-swapping,'' in \emph{Proceedings of the IEEE/CVF International Conference on Computer Vision}, 2023, pp. 7634--7644.

\bibitem{chen2020simswap}
R.~Chen, X.~Chen, B.~Ni, and Y.~Ge, ``Simswap: An efficient framework for high fidelity face swapping,'' in \emph{Proceedings of the 28th ACM International Conference on Multimedia}, 2020, pp. 2003--2011.

\bibitem{li2019faceshifter}
L.~Li, J.~Bao, H.~Yang, D.~Chen, and F.~Wen, ``Faceshifter: Towards high fidelity and occlusion aware face swapping,'' \emph{arXiv preprint arXiv:1912.13457}, 2019.

\bibitem{wang2021hififace}
Y.~Wang, X.~Chen, J.~Zhu, W.~Chu, Y.~Tai, C.~Wang, J.~Li, Y.~Wu, F.~Huang, and R.~Ji, ``Hififace: 3d shape and semantic prior guided high fidelity face swapping,'' \emph{arXiv preprint arXiv:2106.09965}, 2021.

\bibitem{luo2022styleface}
Y.~Luo, J.~Zhu, K.~He, W.~Chu, Y.~Tai, C.~Wang, and J.~Yan, ``Styleface: Towards identity-disentangled face generation on megapixels,'' in \emph{European conference on computer vision}, 2022, pp. 297--312.

\bibitem{xu2022styleswap}
Z.~Xu, H.~Zhou, Z.~Hong, Z.~Liu, J.~Liu, Z.~Guo, J.~Han, J.~Liu, E.~Ding, and J.~Wang, ``Styleswap: Style-based generator empowers robust face swapping,'' in \emph{European Conference on Computer Vision}, 2022, pp. 661--677.

\bibitem{deng2019arcface}
J.~Deng, J.~Guo, N.~Xue, and S.~Zafeiriou, ``Arcface: Additive angular margin loss for deep face recognition,'' in \emph{CVPR}, 2019, pp. 4690--4699.

\bibitem{blanz19993dmm}
V.~Blanz and T.~Vetter, ``A morphable model for the synthesis of 3d faces,'' in \emph{Proceedings of the 26th annual conference on Computer graphics and interactive techniques}, 1999, pp. 187--194.

\bibitem{deng2019accurate}
Y.~Deng, J.~Yang, S.~Xu, D.~Chen, Y.~Jia, and X.~Tong, ``Accurate 3d face reconstruction with weakly-supervised learning: From single image to image set,'' in \emph{Proceedings of the IEEE/CVF Conference on Computer Vision and Pattern Recognition Workshops}, 2019, pp. 0--0.

\bibitem{li20233dswap}
Y.~Li, C.~Ma, Y.~Yan, W.~Zhu, and X.~Yang, ``3d-aware face swapping,'' in \emph{Proceedings of the IEEE/CVF Conference on Computer Vision and Pattern Recognition}, 2023, pp. 12\,705--12\,714.

\bibitem{nirkin2019fsgan}
Y.~Nirkin, Y.~Keller, and T.~Hassner, ``Fsgan: Subject agnostic face swapping and reenactment,'' in \emph{Proceedings of the IEEE/CVF international conference on computer vision}, 2019, pp. 7184--7193.

\bibitem{lee2020maskgan}
C.-H. Lee, Z.~Liu, L.~Wu, and P.~Luo, ``Maskgan: Towards diverse and interactive facial image manipulation,'' in \emph{CVPR}, 2020, pp. 5549--5558.

\bibitem{wang2021faceVid2Vid}
T.-C. Wang, A.~Mallya, and M.-Y. Liu, ``One-shot free-view neural talking-head synthesis for video conferencing,'' in \emph{Proceedings of the IEEE/CVF Conference on Computer Vision and Pattern Recognition}, 2021, pp. 10\,039--10\,049.

\bibitem{yu2021bisenetv2}
C.~Yu, C.~Gao, J.~Wang, G.~Yu, C.~Shen, and N.~Sang, ``Bisenet v2: Bilateral network with guided aggregation for real-time semantic segmentation,'' \emph{International Journal of Computer Vision}, vol. 129, no.~11, pp. 3051--3068, 2021.

\bibitem{karras2020styleGAN2}
T.~Karras, S.~Laine, M.~Aittala, J.~Hellsten, J.~Lehtinen, and T.~Aila, ``Analyzing and improving the image quality of stylegan,'' in \emph{CVPR}, 2020, pp. 8110--8119.

\bibitem{shen2020interfacegan}
Y.~Shen, C.~Yang, X.~Tang, and B.~Zhou, ``Interfacegan: Interpreting the disentangled face representation learned by gans,'' \emph{IEEE transactions on pattern analysis and machine intelligence}, 2020.

\bibitem{richardson2021psp}
E.~Richardson, Y.~Alaluf, O.~Patashnik, Y.~Nitzan, Y.~Azar, S.~Shapiro, and D.~Cohen-Or, ``Encoding in style: a stylegan encoder for image-to-image translation,'' in \emph{CVPR}, 2021, pp. 2287--2296.

\bibitem{wang2022high}
T.~Wang, Y.~Zhang, Y.~Fan, J.~Wang, and Q.~Chen, ``High-fidelity gan inversion for image attribute editing,'' in \emph{CVPR}, 2022, pp. 11\,379--11\,388.

\bibitem{viazovetskyi2020stylegan2}
Y.~Viazovetskyi, V.~Ivashkin, and E.~Kashin, ``Stylegan2 distillation for feed-forward image manipulation,'' in \emph{Computer Vision--ECCV 2020: 16th European Conference, Glasgow, UK, August 23--28, 2020, Proceedings, Part XXII 16}.\hskip 1em plus 0.5em minus 0.4em\relax Springer, 2020, pp. 170--186.

\bibitem{tov2021e4e}
O.~Tov, Y.~Alaluf, Y.~Nitzan, O.~Patashnik, and D.~Cohen-Or, ``Designing an encoder for stylegan image manipulation,'' \emph{ACM Transactions on Graphics (TOG)}, vol.~40, no.~4, pp. 1--14, 2021.

\bibitem{alaluf2021restyle}
Y.~Alaluf, O.~Patashnik, and D.~Cohen-Or, ``Restyle: A residual-based stylegan encoder via iterative refinement,'' in \emph{ICCV}, 2021, pp. 6711--6720.

\bibitem{yao2022FSspace}
X.~Yao, A.~Newson, Y.~Gousseau, and P.~Hellier, ``Feature-style encoder for style-based gan inversion,'' \emph{arXiv e-prints}, pp. arXiv--2202, 2022.

\bibitem{yao2021latent}
{Yao, Xu and Newson, Alasdair and Gousseau, Yann and Hellier, Pierre}, ``A latent transformer for disentangled face editing in images and videos,'' in \emph{Proceedings of the IEEE/CVF international conference on computer vision}, 2021, pp. 13\,789--13\,798.

\bibitem{abdal2019image2stylegan}
R.~Abdal, Y.~Qin, and P.~Wonka, ``Image2stylegan: How to embed images into the stylegan latent space?'' in \emph{ICCV}, 2019, pp. 4432--4441.

\bibitem{abdal2020image2stylegan++}
{Abdal, Rameen and Qin, Yipeng and Wonka, Peter}, ``Image2stylegan++: How to edit the embedded images?'' in \emph{CVPR}, 2020, pp. 8296--8305.

\bibitem{kang2021GANforOORimages}
K.~Kang, S.~Kim, and S.~Cho, ``Gan inversion for out-of-range images with geometric transformations,'' in \emph{ICCV}, 2021, pp. 13\,941--13\,949.

\bibitem{saha2021loho}
R.~Saha, B.~Duke, F.~Shkurti, G.~W. Taylor, and P.~Aarabi, ``Loho: Latent optimization of hairstyles via orthogonalization,'' in \emph{CVPR}, 2021, pp. 1984--1993.

\bibitem{zhu2021barbershop}
P.~Zhu, R.~Abdal, J.~Femiani, and P.~Wonka, ``Barbershop: Gan-based image compositing using segmentation masks,'' \emph{arXiv preprint arXiv:2106.01505}, 2021.

\bibitem{zhu2020indomainGAN}
J.~Zhu, Y.~Shen, D.~Zhao, and B.~Zhou, ``In-domain gan inversion for real image editing,'' in \emph{ECCV}.\hskip 1em plus 0.5em minus 0.4em\relax Springer, 2020, pp. 592--608.

\bibitem{blanz2004exchangingFace}
V.~Blanz, K.~Scherbaum, T.~Vetter, and H.-P. Seidel, ``Exchanging faces in images,'' in \emph{Computer Graphics Forum}, vol.~23, no.~3.\hskip 1em plus 0.5em minus 0.4em\relax Wiley Online Library, 2004, pp. 669--676.

\bibitem{bitouk2008faceSwapping}
D.~Bitouk, N.~Kumar, S.~Dhillon, P.~Belhumeur, and S.~K. Nayar, ``Face swapping: automatically replacing faces in photographs,'' in \emph{ACM SIGGRAPH 2008 papers}, 2008, pp. 1--8.

\bibitem{nirkin2018OnFaceSeg}
Y.~Nirkin, I.~Masi, A.~T. Tuan, T.~Hassner, and G.~Medioni, ``On face segmentation, face swapping, and face perception,'' in \emph{2018 13th IEEE International Conference on Automatic Face \& Gesture Recognition (FG 2018)}.\hskip 1em plus 0.5em minus 0.4em\relax IEEE, 2018, pp. 98--105.

\bibitem{nirkin2022fsganv2}
{Nirkin, Yuval and Keller, Yosi and Hassner, Tal}, ``Fsganv2: Improved subject agnostic face swapping and reenactment,'' \emph{IEEE Transactions on Pattern Analysis and Machine Intelligence}, vol.~45, no.~1, pp. 560--575, 2022.

\bibitem{korshunova2017fastFaceSwap}
I.~Korshunova, W.~Shi, J.~Dambre, and L.~Theis, ``Fast face-swap using convolutional neural networks,'' in \emph{Proceedings of the IEEE international conference on computer vision}, 2017, pp. 3677--3685.

\bibitem{bao2018IPGAN}
J.~Bao, D.~Chen, F.~Wen, H.~Li, and G.~Hua, ``Towards open-set identity preserving face synthesis,'' in \emph{Proceedings of the IEEE conference on computer vision and pattern recognition}, 2018, pp. 6713--6722.

\bibitem{xu2022region}
C.~Xu, J.~Zhang, M.~Hua, Q.~He, Z.~Yi, and Y.~Liu, ``Region-aware face swapping,'' \emph{arXiv preprint arXiv:2203.04564}, 2022.

\bibitem{kim2022smoothswap}
J.~Kim, J.~Lee, and B.-T. Zhang, ``Smooth-swap: A simple enhancement for face-swapping with smoothness,'' in \emph{CVPR}, 2022, pp. 10\,779--10\,788.

\bibitem{ren2023wscswap}
X.~Ren, X.~Chen, P.~Yao, H.-Y. Shum, and B.~Wang, ``Reinforced disentanglement for face swapping without skip connection,'' in \emph{Proceedings of the IEEE/CVF International Conference on Computer Vision}, 2023, pp. 20\,665--20\,675.

\bibitem{zhao2023diffswap}
W.~Zhao, Y.~Rao, W.~Shi, Z.~Liu, J.~Zhou, and J.~Lu, ``Diffswap: High-fidelity and controllable face swapping via 3d-aware masked diffusion,'' in \emph{Proceedings of the IEEE/CVF Conference on Computer Vision and Pattern Recognition}, 2023, pp. 8568--8577.

\bibitem{jiang2023styleipsb}
D.~Jiang, D.~Song, R.~Tong, and M.~Tang, ``Styleipsb: Identity-preserving semantic basis of stylegan for high fidelity face swapping,'' in \emph{Proceedings of the IEEE/CVF Conference on Computer Vision and Pattern Recognition}, 2023, pp. 352--361.

\bibitem{collins2020editingInStyle}
E.~Collins, R.~Bala, B.~Price, and S.~Susstrunk, ``Editing in style: Uncovering the local semantics of gans,'' in \emph{Proceedings of the IEEE/CVF Conference on Computer Vision and Pattern Recognition}, 2020, pp. 5771--5780.

\bibitem{chong2021retrieveInStyle}
M.~J. Chong, W.-S. Chu, A.~Kumar, and D.~Forsyth, ``Retrieve in style: Unsupervised facial feature transfer and retrieval,'' in \emph{Proceedings of the IEEE/CVF International Conference on Computer Vision}, 2021, pp. 3887--3896.

\bibitem{xu2022high}
Y.~Xu, B.~Deng, J.~Wang, Y.~Jing, J.~Pan, and S.~He, ``High-resolution face swapping via latent semantics disentanglement,'' in \emph{Proceedings of the IEEE/CVF Conference on Computer Vision and Pattern Recognition}, 2022, pp. 7642--7651.

\bibitem{park2019semantic}
T.~Park, M.-Y. Liu, T.-C. Wang, and J.-Y. Zhu, ``Semantic image synthesis with spatially-adaptive normalization,'' in \emph{CVPR}, 2019, pp. 2337--2346.

\bibitem{zhu2020sean}
P.~Zhu, R.~Abdal, Y.~Qin, and P.~Wonka, ``Sean: Image synthesis with semantic region-adaptive normalization,'' in \emph{CVPR}, 2020, pp. 5104--5113.

\bibitem{chen2021sofgan}
A.~Chen, R.~Liu, L.~Xie, Z.~Chen, H.~Su, and J.~Yu, ``Sofgan: A portrait image generator with dynamic styling,'' \emph{ACM transactions on graphics}, 2021.

\bibitem{dlib09}
D.~E. King, ``Dlib-ml: A machine learning toolkit,'' \emph{Journal of Machine Learning Research}, vol.~10, pp. 1755--1758, 2009.

\bibitem{karras2019styleGAN}
T.~Karras, S.~Laine, and T.~Aila, ``A style-based generator architecture for generative adversarial networks,'' in \emph{CVPR}, 2019, pp. 4401--4410.

\bibitem{zllrunning2013faceParser}
zllrunning, ``face-parsing.pytorch,'' \url{https://github.com/zllrunning/face-parsing.PyTorch}, 2019.

\bibitem{zhu2020aot}
H.~Zhu, C.~Fu, Q.~Wu, W.~Wu, C.~Qian, and R.~He, ``{AOT}: Appearance optimal transport based identity swapping for forgery detection,'' in \emph{Advances in Neural Information Processing Systems}, 2020, pp. 21\,699--21\,712.

\bibitem{zhang2020cross}
P.~Zhang, B.~Zhang, D.~Chen, L.~Yuan, and F.~Wen, ``Cross-domain correspondence learning for exemplar-based image translation,'' in \emph{CVPR}, 2020, pp. 5143--5153.

\bibitem{zhou2022codeformer}
S.~Zhou, K.~Chan, C.~Li, and C.~C. Loy, ``Towards robust blind face restoration with codebook lookup transformer,'' \emph{Advances in Neural Information Processing Systems}, vol.~35, pp. 30\,599--30\,611, 2022.

\bibitem{roessler2019ffplus}
A.~R\"ossler, D.~Cozzolino, L.~Verdoliva, C.~Riess, J.~Thies, and M.~Nie{\ss}ner, ``Face{F}orensics++: Learning to detect manipulated facial images,'' in \emph{International Conference on Computer Vision (ICCV)}, 2019.

\bibitem{paszke2019pytorch}
A.~Paszke, S.~Gross, F.~Massa, A.~Lerer, J.~Bradbury, G.~Chanan, T.~Killeen, Z.~Lin, N.~Gimelshein, L.~Antiga \emph{et~al.}, ``Pytorch: An imperative style, high-performance deep learning library,'' \emph{Advances in neural information processing systems}, vol.~32, 2019.

\bibitem{kingma2014adam}
D.~P. Kingma and J.~Ba, ``Adam: A method for stochastic optimization,'' \emph{arXiv preprint arXiv:1412.6980}, 2014.

\bibitem{gao2021infoswap}
G.~Gao, H.~Huang, C.~Fu, Z.~Li, and R.~He, ``Information bottleneck disentanglement for identity swapping,'' in \emph{Proceedings of the IEEE/CVF conference on computer vision and pattern recognition}, 2021, pp. 3404--3413.

\bibitem{wang2018cosface}
H.~Wang, Y.~Wang, Z.~Zhou, X.~Ji, D.~Gong, J.~Zhou, Z.~Li, and W.~Liu, ``Cosface: Large margin cosine loss for deep face recognition,'' in \emph{Proceedings of the IEEE conference on computer vision and pattern recognition}, 2018, pp. 5265--5274.

\bibitem{Ruiz2018HopeNet}
N.~Ruiz, E.~Chong, and J.~M. Rehg, ``Fine-grained head pose estimation without keypoints,'' in \emph{The IEEE Conference on Computer Vision and Pattern Recognition (CVPR) Workshops}, June 2018.

\bibitem{heusel2017FID}
M.~Heusel, H.~Ramsauer, T.~Unterthiner, B.~Nessler, and S.~Hochreiter, ``Gans trained by a two time-scale update rule converge to a local nash equilibrium,'' \emph{Advances in neural information processing systems}, vol.~30, 2017.

\bibitem{gu2019mask}
S.~Gu, J.~Bao, H.~Yang, D.~Chen, F.~Wen, and L.~Yuan, ``Mask-guided portrait editing with conditional gans,'' in \emph{CVPR}, 2019, pp. 3436--3445.

\bibitem{wang2004SSIM}
Z.~Wang, A.~C. Bovik, H.~R. Sheikh, and E.~P. Simoncelli, ``Image quality assessment: from error visibility to structural similarity,'' \emph{IEEE transactions on image processing}, vol.~13, no.~4, pp. 600--612, 2004.

\bibitem{zhang2018lpips}
R.~Zhang, P.~Isola, A.~A. Efros, E.~Shechtman, and O.~Wang, ``The unreasonable effectiveness of deep features as a perceptual metric,'' in \emph{CVPR}, 2018, pp. 586--595.

\bibitem{NIPS2012alexnet}
\BIBentryALTinterwordspacing
A.~Krizhevsky, I.~Sutskever, and G.~E. Hinton, ``Imagenet classification with deep convolutional neural networks,'' in \emph{Advances in Neural Information Processing Systems}, F.~Pereira, C.~J.~C. Burges, L.~Bottou, and K.~Q. Weinberger, Eds., vol.~25.\hskip 1em plus 0.5em minus 0.4em\relax Curran Associates, Inc., 2012. [Online]. Available: \url{https://proceedings.neurips.cc/paper/2012/file/c399862d3b9d6b76c8436e924a68c45b-Paper.pdf}
\BIBentrySTDinterwordspacing

\bibitem{krizhevsky2012imagenet}
------, ``Imagenet classification with deep convolutional neural networks,'' \emph{Advances in neural information processing systems}, vol.~25, 2012.

\bibitem{tzaban2022STIT}
R.~Tzaban, R.~Mokady, R.~Gal, A.~H. Bermano, and D.~Cohen-Or, ``Stitch it in time: Gan-based facial editing of real videos,'' \emph{arXiv preprint arXiv:2201.08361}, 2022.

\end{thebibliography}

\newpage

\renewcommand{\figurename}{\textbf{Ap-Fig.}}

\appendices
\section{Training Objectives of the E4S Framework}
\label{sec:lossfuc}
RGI, the core of the proposed E4S, uses reconstruction as the proxy task, where we adopt the commonly used loss functions in the GAN inversion literature.
Besides, the training objective of the proposed face re-coloring network $B_{\psi}$ and face inpainting network $P_{\tau}$ will also be introduced here.




\subsubsection{Loss for the RGI}

\noindent{\textbf{Pixel-wise reconstruction loss.}} 
Defining the input image as $I$ and the corresponding reconstructed image as $\hat{I}$, our pixel-wise reconstruction loss can be expressed as the Mean-Squared-Error (MSE):
\begin{equation}
    \mathcal{L}_{mse} = \left\| \hat{I}-I \right\|_2^2
\end{equation}

\noindent{\textbf{Multi-scale LPIPS loss.}}
Considering that pixel-wise reconstruction loss only cannot lead to a sharp result, inspired by \cite{yao2022FSspace}, we additionally employ the multi-scale LPIPS~\cite{zhang2018lpips} loss for a more sharp reconstructed result. The loss term is expressed as follows:
\begin{equation}
    \mathcal{L}_{ms\_ lpips}=\sum_{s} \left\|  \mathbf{V}(\lfloor\hat{I}\rfloor_{s})-\mathbf{V}(\lfloor I\rfloor_{s}) \right\|_2^2 \ ,
    \small \forall s \in\{256,512,1024\},
\end{equation}
where $\mathbf{V}$ represents the AlexNet~\cite{NIPS2012alexnet} feature extractor pre-trained on ImageNet~\cite{krizhevsky2012imagenet} and $\lfloor\hat{I}\rfloor_{s}$ represents the downsized input in resolution of $s$.
This multi-scale perceptual loss allows the style codes to contain perceptual similarities at different levels.

\noindent{\textbf{Multi-scale face inversion loss.}}
The previous work PSP~\cite{richardson2021psp} proposes an ID loss to maintain the input identity.
Specifically, it leverages a pre-trained face recognition network, which encourages the 
cosine similarity between the input and the reconstructed face to be maximized.
Besides, \cite{yao2022FSspace} further 
improve the ID loss with multi-scale form, 
which computing similarities in different feature levels. 
Following these two work, we define our multi-scale ID loss term as:
\begin{equation}
    \mathcal{L}_{ms\_ id} = \sum_{i=1}^{5}\left(1-\langle R_i(I),R_i(\hat{I})\rangle\right),
\end{equation}
where $\langle\cdot\rangle$ denotes the cosine similarity and $R$ is the pre-trained ArcFace~\cite{deng2019arcface} model.

Furthermore, following \cite{yao2022FSspace}, we apply a multi-scale face parsing loss for a more accurate parsing, which can be expressed as:
\begin{equation}
    \mathcal{L}_{ms\_ parsing} = \sum_{i=1}^{5}\left(1-\langle P_i(I),P_i(\hat{I})\rangle\right),
\end{equation}
where $P$ is the pre-trained face parser~\cite{lee2020maskgan}.

Finally, our reconstruction loss $\mathcal{L}_{recon}$ can be expressed as:
\begin{equation}
\label{eqn:recon}
\mathcal{L}_{recon} = \mathcal{L}_{mse} + \lambda_1\mathcal{L}_{ms\_ lpips} + \lambda_2\mathcal{L}_{ms\_ id} + \lambda_3\mathcal{L}_{ms\_ parsing},
\end{equation}
where $\lambda_1$, $\lambda_2$, and $\lambda_3$ are trade-off hyperparameters.

\noindent{\textbf{Adversarial loss.}}
Besides of the reconstruction loss $\mathcal{L}_{recon}$, 
we additionally use adversarial training to help improve the final image quality, which is expressed as:
\begin{equation}
    \mathcal{L}_{adv}= \mathbb{E}[1-\log D(\hat{I})] + \mathbb{E}[\log D(I)],
\end{equation}
where $D$ is initialized with the pre-trained StyleGAN discriminator. 
Finally, the overall loss function of our RGI is defined as:
\begin{equation}
    \mathcal{L} = \mathcal{L}_{recon} + \lambda_{adv}\mathcal{L}_{adv},
\end{equation}
where the hyperparameters $\lambda_1,\lambda_2,\lambda_3$, and $\lambda_{adv}$ are set as 0.8, 0.1, 0.1, and 0.01, respectively in all experiments.

\subsubsection{Loss for the Face Re-coloring Network $B_{\psi}$}

\textbf{Reconstruction Loss.}
During training, given $I'_{\rm{A}}$ and and $I'_{\rm{AG}}$, the re-coloring network learns to predict the re-colored $I_{\rm{rec}}$.
The predicted $I_{\rm{rec}}$ should be fully consistent with the $I_{\rm{A}}$ (which equals to the flipped $I_{\rm{A}}'$).
To this end, we use a reconstruction loss consisting of an L2 loss and an LPIPS loss as the training objective:
\begin{equation}
    \mathcal{L_{\psi,\rm{rec}}} = \lambda_{\psi, 1}\|I_{\rm{A}} - I_{\rm{rec}}\|^2_2+ \lambda_{\psi, 2}\| {\rm{V}}(I_{\rm{A}}) - {\rm{V}}(I_{\rm{rec}}) \|^2_2,
\end{equation}
where $\lambda_{\psi, 1}=1$ and $\lambda_{\psi, 2}=1$ can yield reasonable results without grid search in our experiments.

\subsubsection{Details of Training the Face Inpainting Network $P_{\tau}$}

\noindent\textbf{Training Settings and Loss Function.}
Similar to training re-coloring network $B_{\psi}$, training inpainting network $P_{\tau}$ requires paired data which is hard to collect.
Therefore, we train $P_{\tau}$ in a self-supervised scheme.
Given a face $I$, we generate random mismatch masks around the face contour and edit these regions based on the editing ability of our RGI method. The edited face $I_{\rm{edit}}$ have an inconsistent face shape with the original one.
Then we adopt $I_{\rm{edit}}$ as the training input and $I$ as the ground-truth supervision with a reconstruction loss, enforcing the inpainting network to predict a shape-consistent face result $I_{\rm{inp}}$ under the guidance of the aforementioned random mismatch mask:
\begin{equation}
    \mathcal{L_{\tau,\rm{inp}}} = \lambda_{\tau, 1}\|I - I_{\rm{inp}}\|^2_2+ \lambda_{\tau, 2}\| {\rm{V}}(I) - {\rm{V}}(I_{\rm{inp}}) \|^2_2,
\end{equation}
where $\lambda_{\tau, 1}=1$ and $\lambda_{\tau, 2}=5$ in our experiments.


\begin{figure*}[t]
\centering
\setkeys{Gin}{width=\linewidth}
\captionsetup{belowskip=-15pt}
\begin{subfigure}{\linewidth}
\includegraphics[width=1.0\textwidth]{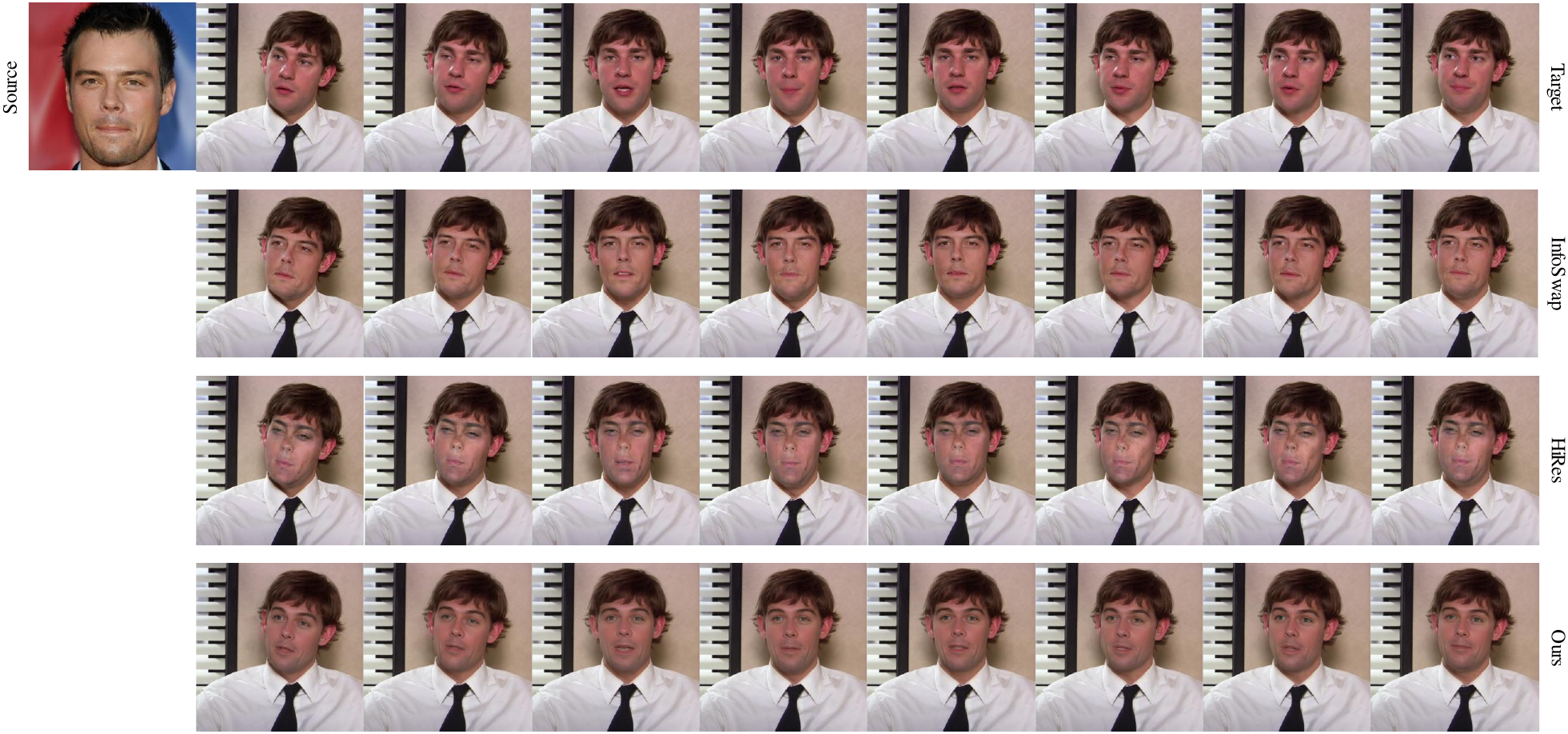}
\vspace{10pt}
\end{subfigure}
\begin{subfigure}{\linewidth}
\includegraphics[width=1.0\textwidth]{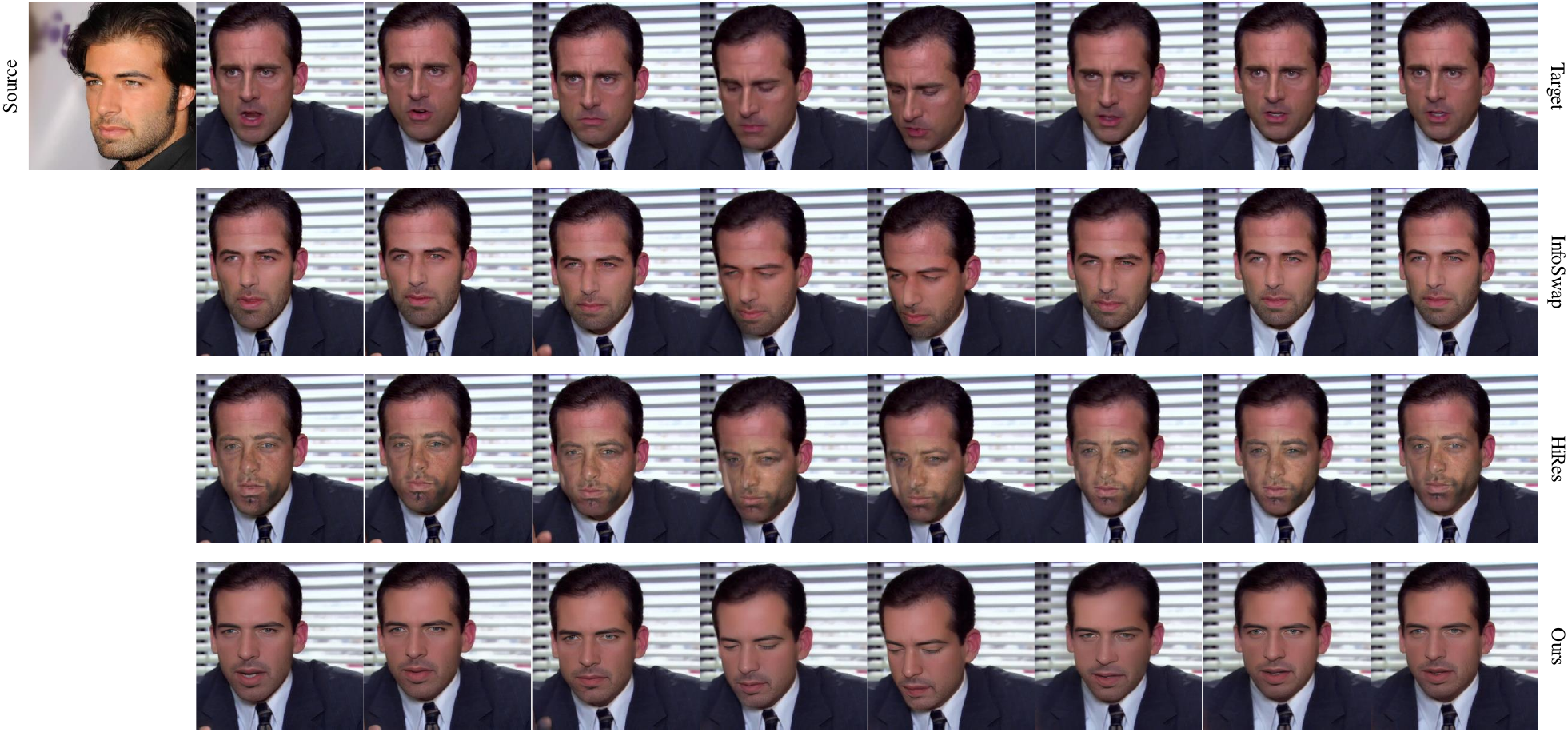}
\end{subfigure}
\caption{Video face swapping comparisons of our results with FaceShifter~\cite{li2019faceshifter} and HiRes~\cite{xu2022high}. Our method shows the better capability
of source identity transferring and target attribute preservation (e.g., pose, expression, wink, lighting). Their visual quality and temporal consistency
also inferior to our method.}
\label{fig:videoswap}
\end{figure*}
\begin{figure*}[ht]
\centering
\setkeys{Gin}{width=\linewidth}
\captionsetup{belowskip=-10pt}
\begin{subfigure}{\linewidth}
\includegraphics[width=1.0\textwidth]{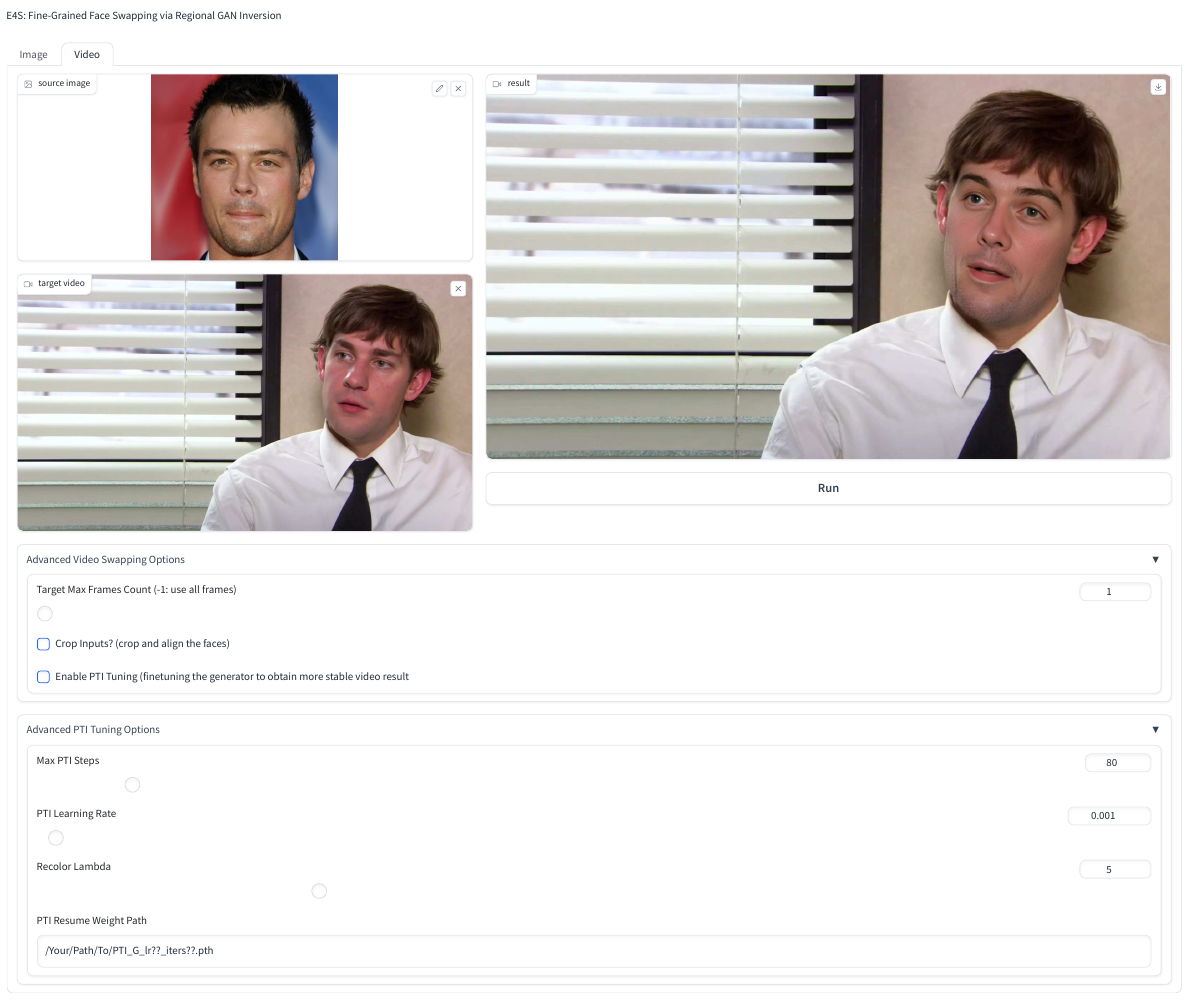} 
\end{subfigure}
\caption{A screenshot of our interactive image and video face swapping system. The button in the upper left corner can choose whether to swap face in images or videos.}
    \label{fig:screen}
\end{figure*}

\section{Video Face Swapping Results}
\label{sec:videoface}
The proposed E4S framework can be used to conduct video face
swapping. Specifically, following STIT~\cite{tzaban2022STIT}, we first crop and align the source
image and the target video in advance, thus resulting in the source face S and an n-frame target face video $\{T_i\}_{i=1}^n$. 
Then, as described in Sec.
\textcolor{red}{III-A}
in our main paper, we can perform face re-enactment on the source S and make it have a similar pose and expression as each target frame. In this way, we can obtain a sequence of driven video $\{D_i\}_{i=1}^n$. 
Next, we can perform face swapping between each driven and target pair $\{(D_i,T_i)\}_{i=1}^n$.

Nonetheless, since our RGI is trained on images rather than videos, we find the above-mentioned video swapping results would have some temporal inconsistency.
To deal with this, we turn to fine-tune the generator of our RGI on all the driven frames, where the $\mathcal{L}_{recon}$ is the loss function. The parameters are updated 200 times for each frame, where the learning rate is $10^{-3}$. After finetuning, we still adopt the frame-by-frame swapping strategy to obtain a temporally consistent result.

We compare our results with FaceShifter~\cite{li2019faceshifter} and HiRes~\cite{xu2022high}
in Ap-Fig.~\ref{fig:videoswap}, where the whole video is on our project page.
HiRes struggles to generate a wink in the swapped video. Besides, FaceShifter sometimes fails to transfer the source identity. As contrast,  our results show better visual quality and temporal consistency.


\section{User-interface System For Image and Video Face Swapping}
\label{sec:user}
We develop a user-interface system to perform image and video face swapping. A screenshot is shown in Ap-Fig.~\ref{fig:screen}.
Given a source image and a target image or video, our user-interface system can produce a high-quality swapped image or video. An interactive face swapping  demo can also be found on our project page.





 





\end{document}